\pdfoutput=1

\documentclass[11pt]{article}

\usepackage{acl}
\bibstyle{acl_natbib}

\usepackage{times}
\usepackage{latexsym}
\usepackage{graphicx}
\usepackage{booktabs}
\usepackage{multirow}  
\usepackage{subcaption}
\usepackage{caption}
\usepackage{tabularx}
\usepackage{wrapfig}
\usepackage[T1]{fontenc}

\usepackage[utf8]{inputenc}

\usepackage{microtype}

\usepackage{inconsolata}

\title{Towards Region-aware Bias Evaluation Metrics}

\author{Angana Borah\textsuperscript{{\normalfont 1}} Aparna Garimella\textsuperscript{{\normalfont 2}} {\bf Rada Mihalcea\textsuperscript{{\normalfont 1}}} \\  
  \textsuperscript{1}University of Michigan, Ann Arbor, USA,
  \textsuperscript{2}Adobe Research, India \\
  \\
  {\tt anganab@umich.edu} \hspace{0.5cm} 
  {\tt garimell@adobe.com} \hspace{0.5cm} 
  {\tt mihalcea@umich.edu}
  }

\begin{document}
{\makeatletter\acl@finalcopytrue
  \maketitle
}
\begin{abstract}
When exposed to human-generated data, language models are known to learn and amplify societal biases. While previous works introduced benchmarks that can be used to assess the bias in these models, they rely on assumptions that may not be universally true. For instance, a gender bias dimension commonly used by these metrics is that of {\it family--career}, but this may not be the only common bias in certain regions of the world. In this paper, we identify topical differences in gender bias across different regions and propose a region-aware bottom-up approach for bias assessment. Our proposed approach uses gender-aligned topics for a given region and identifies gender bias dimensions in the form of topic pairs that are likely to capture gender societal biases. Several of our proposed bias topic pairs are on par with human perception of gender biases in these regions in comparison to the existing ones, and we also identify new pairs that are more aligned than the existing ones. In addition, we use our region-aware bias topic pairs in a Word Embedding Association Test (WEAT)-based evaluation metric to test for gender biases across different regions in different data domains. We also find that LLMs have a higher alignment to bias pairs for highly-represented regions showing the importance of region-aware bias evaluation metric.
\end{abstract}

\section{Introduction}



Human bias refers to the tendency of prejudice or preference towards a certain group or an individual and can reflect social stereotypes with respect to gender, age, race, religion, and so on. 

Bias in machine learning refers to prior information which is a necessary prerequisite for intelligence \cite{bishop2006pattern}. However, biases can be problematic when prior information is derived from {\it harmful precedents} like prejudices and social stereotypes. Early work in detecting biases includes the Word Embedding Association Test (WEAT) \cite{caliskan2017semantics} and the Sentence Encoder Association Test (SEAT) \cite{may-etal-2019-measuring}. WEAT is inspired by the Implicit Association Test (IAT) \cite{greenwald1998measuring} in psychology, which gauges people's propensity to unconsciously link particular characteristics—like {\it family} versus {\it career}—with specific target groups—like female (F) versus male (M). 
WEAT measures the distances between target and attribute word sets in word embeddings using dimensions\footnote{We use `topic pairs' and `topic dimensions' interchangeably.} similar to those used in IAT.  

Biases toward or against a group can vary across different regions due to the influence of an individual's culture and demographics \cite{grimm1999cross, kiritchenko2018examining, garimella-etal-2022-demographic}. Psychological studies and experiments that demonstrate human stereotypes vary by continental regions \cite{damann2023persistence, nationalgeographic} and even larger concepts like western and eastern worlds \cite{markus2003models, jiang2019cultural} serve as an inspiration for the use of regions to determine differences across cultures. However, existing bias evaluation metrics like WEAT and SEAT follow a ``one-size-fits-all'' approach to detect biases across different regions. 
As biases can be very diverse depending on the demographic lens, a fixed or a small set of dimensions (such as family--career, math--arts) may not be able to cover all the possible biases in the society.
In this paper, we address two main research questions about gender bias: (1) Is it possible to use current NLP techniques to automatically identify gender bias characteristics (such as family, career) specific to various regions? (2) How do these gender dimensions compare to the current generic dimensions  included in WEAT/SEAT?

Our paper makes four main contributions: 
\begin{enumerate}
    \item  An automatic method to uncover gender bias dimensions in various regions that uses (a) topic modeling to identify dominant topics aligning with the F/M groups for different regions, and (b) an embedding-based approach to identify F-M topic pairs for different regions that can be viewed as gender bias dimensions in those regions.
    \item An IAT-style test to assess our predicted gender bias dimensions with human subjects. To the best of our knowledge, this is the first study to use a data-driven, bottom-up method to evaluate bias dimensions across regional boundaries. 
    \item A WEAT-based evaluation setup using our region-aware topic pairs to evaluate gender biases in different data domains  (Reddit and UN General Debates) across regions.
    \item An analysis of how well our predicted bias dimensions align with those of custom LLMs. We consider several LLMs that include open-source models like \texttt{Llama-3-8b} and \texttt{Mistral-7b-Instruct}; as well as closed-source models such as \texttt{GPT-4}, \texttt{Gemini-Pro} and \texttt{Claude-3-Sonnet}. 
\end{enumerate}







\section{Data}
We use GeoWAC \cite{dunn2020geographically}, a geographically balanced corpus that consists of web pages from Common Crawl. Language samples are geo-located using country-specific domains, such as an $.in$ domain suggesting Indian origin \cite{dunn2020mapping}. The GeoWAC's English corpus spans 150 countries. We select the top three countries with the most examples per region: Asia, Africa, Europe, North America, and Oceania as in \cite{garimella-etal-2022-demographic}. We randomly choose 282,000 examples (after pre-processing) for each region, with 94,000 examples belonging to each country within the regions. Dataset details are included in Appendix~\ref{geowac_data}.

\section{Variations in Gender Bias Tests Across Regions}
We start by investigating the differences in existing gender bias tests across different regions using WEAT. WEAT takes in {\it target words} such as male names and female names, to indicate a specific group, and {\it attribute words} that can be associated with the \textit{target words}, such as {\it math} and {\it art}. It computes bias by finding the cosine distance between the embeddings of the target and attribute words. 
We compute WEAT scores using word2vec embeddings \cite{mikolov2013efficient} trained on the five regions separately. Table~\ref{weat_cont} shows the region-wise scores for the three gender tests in WEAT. 
\definecolor{PastelPurple}{RGB}{216, 191, 216}  
\begin{table}[t]
\centering
\scalebox{0.75}{
\begin{tabular}{p{5cm}|p{1.5cm}|r}
\toprule
\textbf{\textsc{Target words - Attribute words}} & \textbf{\textsc{Region}} & \textbf{\textsc{WEAT}} \\ 
\midrule
  & Africa & 1.798 \\ 
& Asia & 1.508 \\ 
  career vs family  - Male names vs Female names   & North America & \colorbox{PastelPurple}{1.885} \\ 
 & Europe & 1.610 \\ 
 & Oceania & 1.727 \\ 
 \midrule

& Africa & \colorbox{PastelPurple}{1.429} \\ 
& Asia & 1.187 \\ 
Math vs Arts - Male terms vs Female terms  & North America &  0.703 \\ 
 & Europe & 0.334 \\ 
 & Oceania & 1.158 \\ 
 \midrule

 & Africa & \colorbox{PastelPurple}{1.247} \\ 
  & Asia & 0.330 \\ 
  Science vs Arts - Male terms vs Female terms & North America &  0.036 \\ 
 & Europe & -0.655 \\ 
 & Oceania & 0.725 \\ \bottomrule 

\end{tabular}
}
\caption{Region-wise WEAT scores using word2vec.}
\label{weat_cont}
\vspace{-0.2cm}
\end{table}

Although we see a positive bias for most gender bias dimensions, the scores vary across regions. 
For example, the highest scoring regions vary for the target words-attribute words groups. For {\it family--career} dimension, North America shows the highest bias, however for {\it math--arts} and {\it math--science} dimensions, Africa shows the highest bias. 
Europe has a negative bias on {\it science--arts} (indicating a stronger F-science and M-arts association). 

These results provide preliminary support to our hypothesis that gender bias dimensions vary across regions, thus propelling a need to come up with further bias measurement dimensions to better capture gender biases in these regions in addition to the existing generic ones in WEAT.

\section{A Method to Automatically Detect Bias Dimensions Across Regions}
\begin{figure*}
    \centering
    \includegraphics[width=1\linewidth]{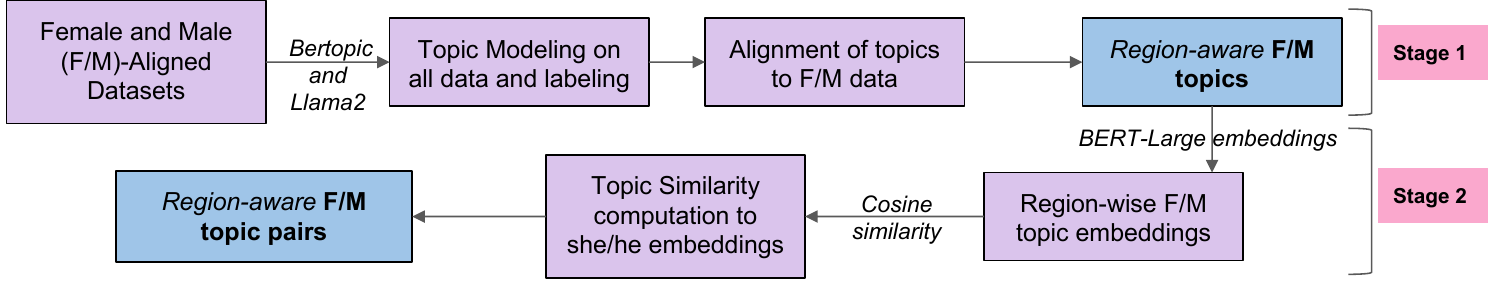}
    \caption{Methodology Pipeline: Stage 1 refers to the extraction of region-aware gender topics using topic modeling, Stage 2 refers to extraction of region-aware gender topic pairs using an embedding based approach}
    \label{fig:ppl}
\end{figure*}

Building upon our WEAT findings, we propose a two-stage approach to automatically detect region-aware bias dimensions that likely capture the biases in specific regions in a bottom-up manner.
In the first stage, we utilize topic modeling to identify prominent topics in each region. In the second stage, we  use an embedding-based approach to find pairs of topics among those identified in the first stage that are likely to represent prominent gender bias dimensions in each region. 
Fig~\ref{fig:ppl} shows the pipeline of our methodology. 


\subsection{Identifying Region-wise Bias Topics}
We use topic modeling to identify dominant topics in the male and female examples in each region.

We first build F(emale)- and M(ale)-aligned datasets  using the examples from GeoWAC for each region.  
We use the 52 pairs of gender-defined words that are non-stereotypically F/M (e.g., wife, brother, see Appendix~\ref{bolukbasilist}) from \cite{bolukbasi2016man}, and find examples that contain these words.
These datasets are used to find gender-aligned topics from GeoWAC. 
The dataset statistics 
are specified in Table~\ref{Dataset_Cont} in Appendix~\ref{fm_appendix}. 

We then use topic modeling to identify dominant topics in the male and female examples in each region.
We use \texttt{Bertopic} \cite{grootendorst2022bertopic}, which identifies an optimal number of topics $n$ for a given dataset (see Appendix~\ref{bertopic} for implementation details). We further refine the resulting topics using \texttt{Llama2} \cite{touvron2023llama} to label and better understand the topic clusters identified by \texttt{Bertopic}. The prompting mechanism for \texttt{Llama2} is provided in Appendix~\ref{prompt}. 

We next compute the alignment of the topics to either of the F/M groups. We first compute the topic distribution of a data point, which gives the probability $p_{it}$ of an example $i$ belonging to each topic $t$. For a topic $t$, we take n examples that dominantly belong to topic $t$: $i_1, i_2, ...., i_n$. If m out of n data points belong to the F group in the F-M dataset, and the other (n - m) belongs to the M group, we compute the average of topic probabilities for both groups separately: $p_{Ft} = \frac{(p_{i_{1}t} + p_{i_{2}t} + ...... + p_{i_{m}t})}{m}$ and $p_{Mt} = \frac{(p_{i_{m+1}t} + p_{i_{m+2}t} + ...... + p_{i_{n}t})}{(n-m)}$, where $p_{Ft}$ and $p_{Mt}$ refer to the average probability by which a topic dominantly belongs to the F and M groups respectively. If $p_{Ft} > p_{Mt}$, we say the topic is a {\it bias topic} that aligns with the F group and vice-versa. 

\subsection{Finding Topic Pairs as Region-wise Bias Dimension Indicators}
We use an embedding-based approach to identify F-M topic pairs from the pool of topics identified in the previous stage, to generate topic pairs (bias dimensions) that are comparable to IAT/WEAT pairs.

We use BERT-large (\texttt{stsb-bert-large}) from SpaCy's \cite{spacy2} \texttt{sentencebert} library to extract contextual embeddings for topic words extracted from the GeoWAC dataset for each region. 
For a topic $t$ consisting of topic words $w_1, .. w_n$, the topic embedding is given by the average of embeddings of the top ten topic words in that topic.

We identify topic pairs from the embeddings taking inspiration from \cite{bolukbasi2016man}: let the embeddings of the words $she$ and $he$ be $E_{she}$ and $E_{he}$ respectively. The embedding of a topic $t_i$ be $E_{t_i}$. A female topic $F_{t_i}$ and a male topic $M_{t_j}$ are a topic pair if: $cos(E_{F_{t_i}}, E_{she}) \sim \ cos(E_{M_{t_j}}, E_{he})$ and/or $cos(E_{F_{t_i}}, E_{he}) \sim \ cos(E_{M_{t_j}}, E_{she})$, where $cos(i, j)$ refers to the cosine similarity between embeddings $i$ and $j$, given by $cos(i,j) = \frac{i,j}{||i||||j||}$. The threshold for the difference between the cosine similarities we consider for two topics to be a pair is 0.01, i.e., two topics $(t1, t2)$ are considered a pair if the difference of cosine similarities cos($t1$, $she$)/cos($t1$, $he$) and cos($t2$, $he$)/cos($t2$, $she$) respectively is $< 0.01$. We manually choose 0.01 since differences close to 0.01 are almost $=0$. 

\begin{figure*}
    \centering
    \includegraphics[width=0.9\linewidth]{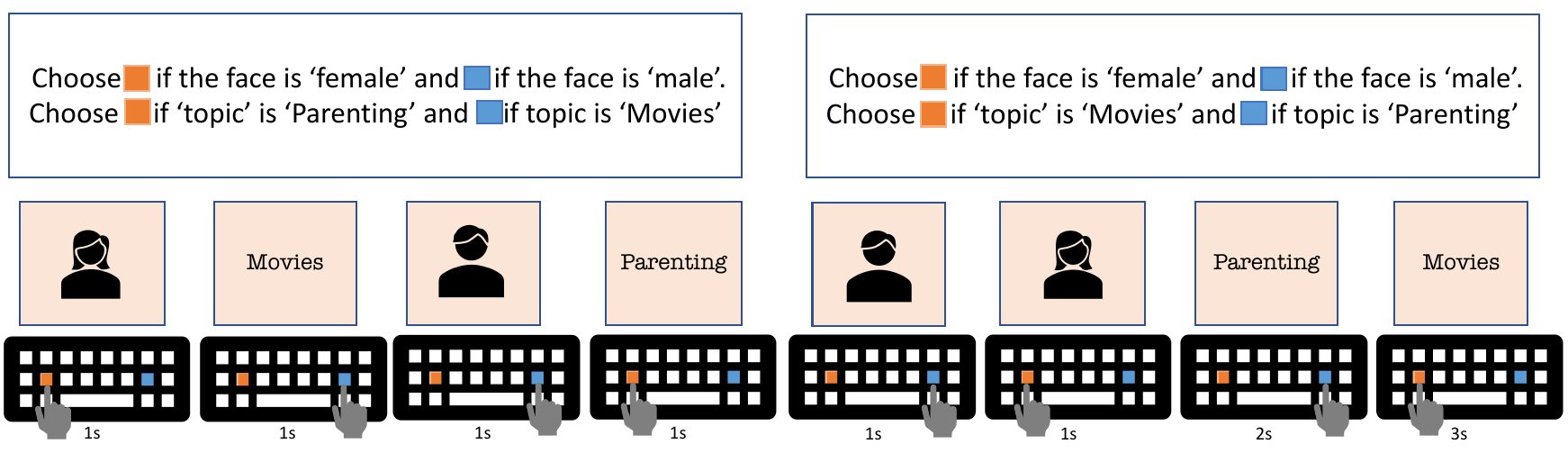}
    \caption{IAT-style test with region-aware topic pairs for human validation. The above example shows the user implicitly associates female to \textit{parenting} and male to \textit{movies}: When guidelines are reversed, they take longer time. Note that we randomize the order of tests for participants to ensure initial pairing bias is accounted for. Also, we have several pages showing faces and topics for each guideline.}
    \label{fig:iat_region}
\end{figure*}
\subsection{Human Validation Setup}
We design an IAT-style test to validate our topic pairs with annotators from different regions. We recruit six annotators from each region controlled by gender (three female and three male). In addition to our topics, we also test for existing WEAT dimensions relating to gender, namely \textit{family--career, math--arts, and science--arts}. For each region, we validate all the region-aware topic pairs using the assistance of our annotators.

As done in IAT, to verify a topic pair, we show the topic names and male/female faces to our annotators along with a set of guidelines. As shown in Fig~\ref{fig:iat_region}, each topic pair test form contains two tasks. 
First, the annotators have to press one key for a female face $f$ and a female topic $T_f$ and another key for a male face $m$ with a male topic $T_m$, timing responses as $r_1$ and $r_2$. 
In the reverse task, they pair $T_m$ with $f$ and $T_f$ with $m$, timing these as $r_3$ and $r_4$. We average $r_1$ and $r_2$ for the `un-reversed' case and $r_3$ and $r_4$ for the `reversed' case. The annotators' implicit association of a gender to a topic may influence their response time. A lower response time suggests easier recollection of the guidelines and potential implicit gender-topic associations, and thus lower bias with respect to these topics. We also varied the test order for different annotators to avoid initial pairing bias. 
We conduct the survey with six annotators each from Africa, Asia, Europe, and North America, also including a {\it family-career} topic pair, a standard WEAT bias dimension. We provide screenshots of our annotation framework in Appendix~\ref{sec:screenshots}.  
 

\subsection{Results: Bias Dimensions across Regions}
\noindent

\subsubsection{Region-wise Bias Topics} Table~\ref{continent_toptopics} displays the top topics based on $u_{mass}$ \cite{mimno-etal-2011-optimizing} coherence for each region.

\begin{table}[t]  
\scriptsize  
\centering  
\begin{tabularx}{\columnwidth}{l|X|X}  
\toprule  
\textbf{\textsc{Region}} & \textbf{\textsc{Female}} & \textbf{\textsc{Male}} \\ \midrule  
  
Africa & Credit cards and finances, Royalty and Media, Trading strategies and market analysis, Dating and relationships guides, Parenting and family relationships & Fashion and Lifestyle, Male enhancement and sexual health, Nollywood actresses and movies, Nigerian politics and government, Essay writing and research \\ \midrule  
  
Asia & Hobbies and Interests, Healthy eating habits for children, Social media platforms, Royal wedding plans, Online Dating and Chatting & DC comic characters, Mobile Application, Phillippine Politics and Government, Sports and Soccer, Career \\ \midrule  
  
Europe & Pets and animal care, Fashion and Style, Education, Obituaries and Genealogy, Luxury sailing & Political developments in Northern Ireland, Christian Theology and Practice, Crime and murder investigation, EU Referendum and Ministerial Positions, Criminal Justice System \\ \midrule  
  
North America & Pets, Cooking: culinary delights and chef recipes, Fashion and style, Family dynamics and relationships, Reading and fiction & Civil War and history, Middle East conflict and political tensions, Movies and filmmaking, Political leadership and party dynamics in Bermuda, Rock Music and songwriting \\ \midrule  
  
Oceania & Cooking and culinary delights, Romance, Weight loss and nutrition for women, Water travel experience, Woodworking plans and projects & Harry Potter adventures, Art and Photography, Superheroes and their Universes, Music recording and Artists, Football in Vanuatu \\ \bottomrule  
  
\end{tabularx}  
\caption{Top five topics for F and M for each region, extracted using Bertopic and Llama2.}  
\label{continent_toptopics} 
\vspace{-0.15cm}
\end{table}

Several topics are exclusive to certain regions. Some topics like {\it family} and {\it parenting}; {\it cooking}; {\it pets} and {\it animal care} are common across some regions for F. Similarly we have {\it movies}; {\it politics} and {\it government}; {\it sports}; and {\it music} for M. Finally, there are differences between regions in terms of {\it education}, {\it reading}, and {\it research} (F-Europe, NA, and M-Africa), and {\it fashion} and {\it lifestyle} (F-Europe, NA, and M-Africa). Some other popular topics across regions are {\it religion and spirituality}, {\it Christian theology} in M; {\it obituaries and genealogy}, {\it online dating}, {\it travel}, and {\it sailing} in F (see Appendix~\ref{topiclist_appendix} for a comprehensive list of topics). We provide an example of topic clusters in Appendix~\ref{sec:topic_example}. 

\noindent
\subsubsection{Region-wise Bias Dimensions} Table~\ref{continent_topicpairs} shows the top five topic pairs per region, chosen based on the $u_{mass}$ score from the top 10 topics each for F and M from the topic modeling scheme. As expected, topic pairs differ by region, and we also note new topic pairs that do not appear in the WEAT tests. Among the top ones, there are recurring topics in F such as {\it dating and marriage}, {\it family and relationships}, {\it luxury sailing}, and {\it education}, whereas in M, we have {\it politics}, {\it religion}, {\it sports}, and {\it movies}. 
These region-specific pairs may supplement generic tests to detect regional biases. 


\noindent

\begin{table}[!ht]
\scriptsize
\centering
\begin{tabular}{p{1cm}|p{5.8cm}}
\toprule
\textbf{\textsc{Region}} & \textbf{\textsc{F-M topic pair}}\\ \midrule
\multirow{5}{*}{Africa} & Parenting and family relationships-Nollywood Actress and Movies (P1) \\ 
& Marriage and relationships - Sports and Football (P2) \\ 
& Womens' lives and successes - Fashion and Lifestyle (P3) \\ 
& Music - Social Media (P4)\\ 
& Dating and relationships advice - Religious and Spiritual growth (P5) \\ \midrule   

\multirow{5}{*}{Asia} & Hotel royalty - Political leadership in India (P1)\\ 
& Healthy eating habits for children - Sports and Soccer (P2) \\ 
& Royal wedding plans - Social Media platforms for video sharing (P3) \\ 
& Royal wedding plans - Religious devotion and spirituality (P4) \\ 
& Marriage - Bollywood actors and films (P5) \\ \midrule 

\multirow{5}{*}{Europe} & Education - Music (P1) \\ 
& Comfortable hotels - Political decision and impact on society (P2) \\ 
& Luxury sailing - UK Government Taxation policies (P3) \\
& Obituaries and Genealogy - Christian Theology and Practice (P4) \\ 
& Fashion and style - Christian theology and practice (P5) \\ \midrule

 & Online Dating for Singles - Religion and Spirituality (P1) \\ 
North & Fashion and Style - Reproductive Health (P2) \\ 
America & Education and achievements - Reinsurance and capital markets (P3) \\ 
& Family dynamics and relationships - Nike shoes and fashion (P4) \\ 
& Reading and fiction - Cape Cod news (P5) \\ \midrule  

\multirow{5}{*}{Oceania} & Family relationships - Religious beliefs and figures (P1) \\
& Woodworking plans and projects - Music record and Artists (P2) \\
& Weight loss and nutrition for women - Building and designing boats (P3) \\ 
& Exercises for hormone development - Superheroes and their Universes (P4) \\ 
& Kids' furniture and decor - Building and designing boats (P5) \\ 
\bottomrule  
\end{tabular}
\caption{\label{continent_topicpairs} \centering
Top five region-aware topic pairs for F and M for each region using en embedding-based approach.
}
\end{table}

\begin{figure*}[ht]
    \centering  
    \begin{subfigure}{0.47\textwidth}
        \includegraphics[width=\textwidth]{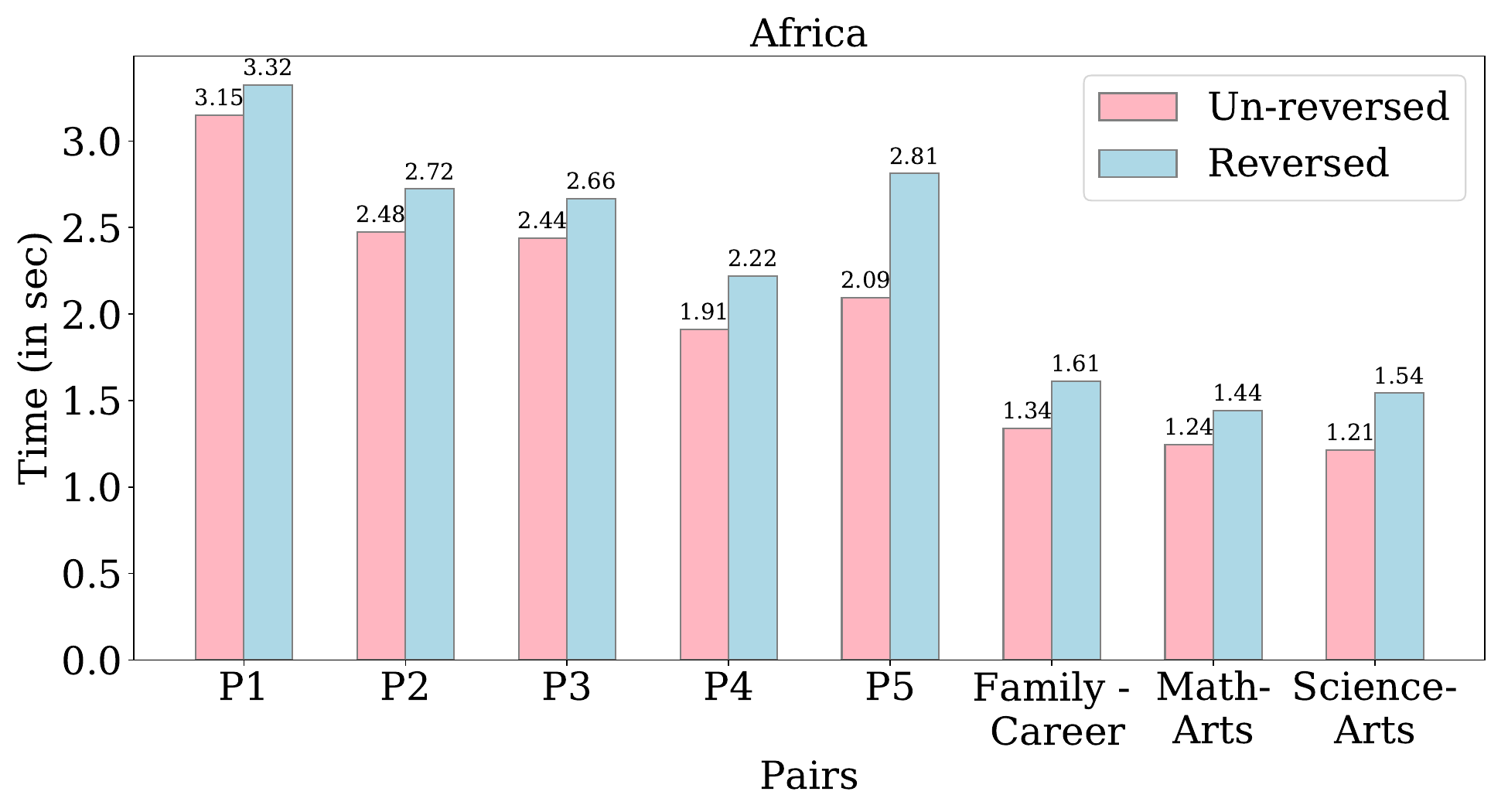}  
    \end{subfigure}  
    \begin{subfigure}{0.47\textwidth}
        \includegraphics[width=\textwidth]{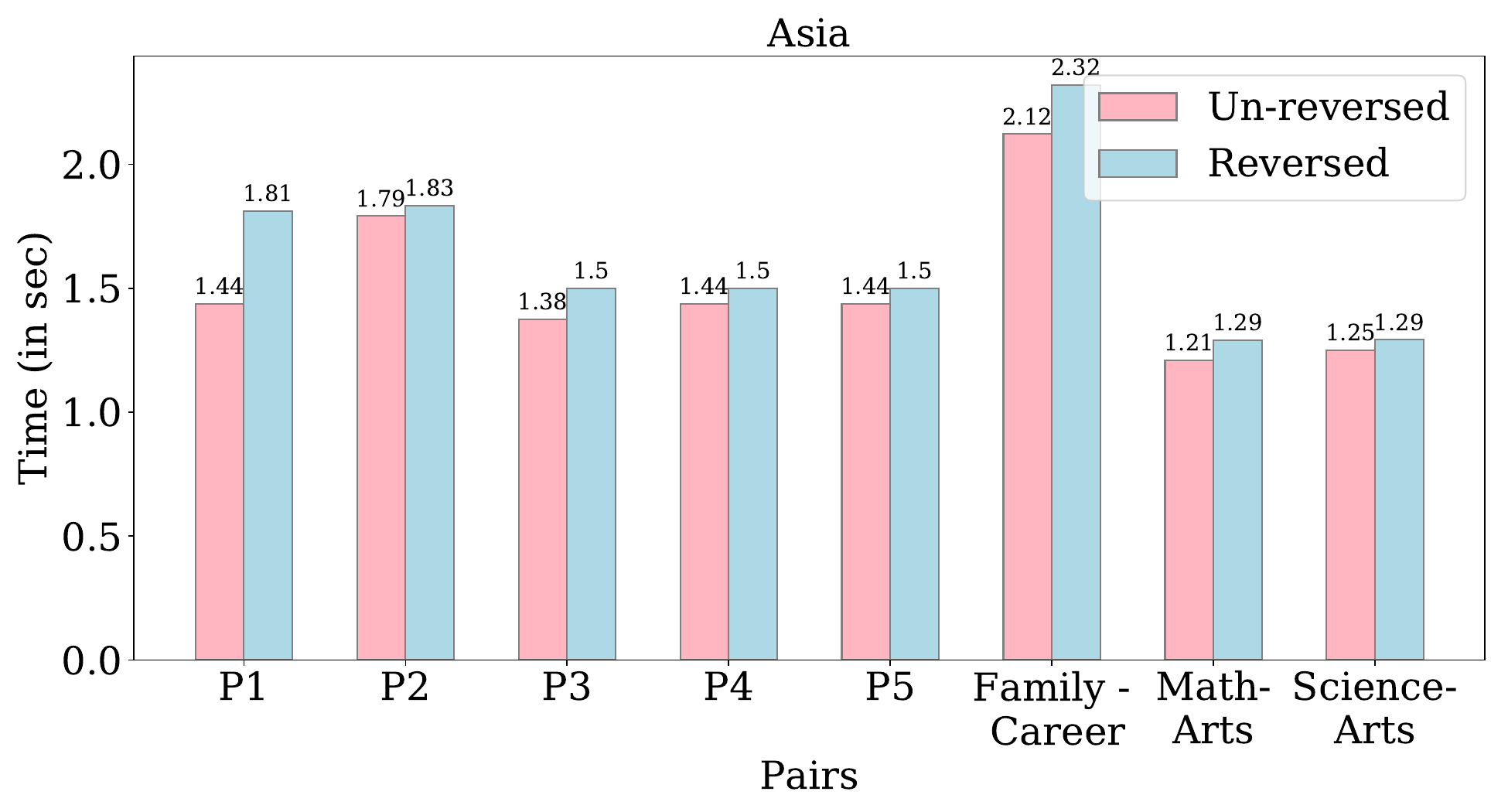}  
    \end{subfigure}  
      
    \vspace{0cm}  
      
    \begin{subfigure}{0.47\textwidth}
        \centering  
        \includegraphics[width=\textwidth]{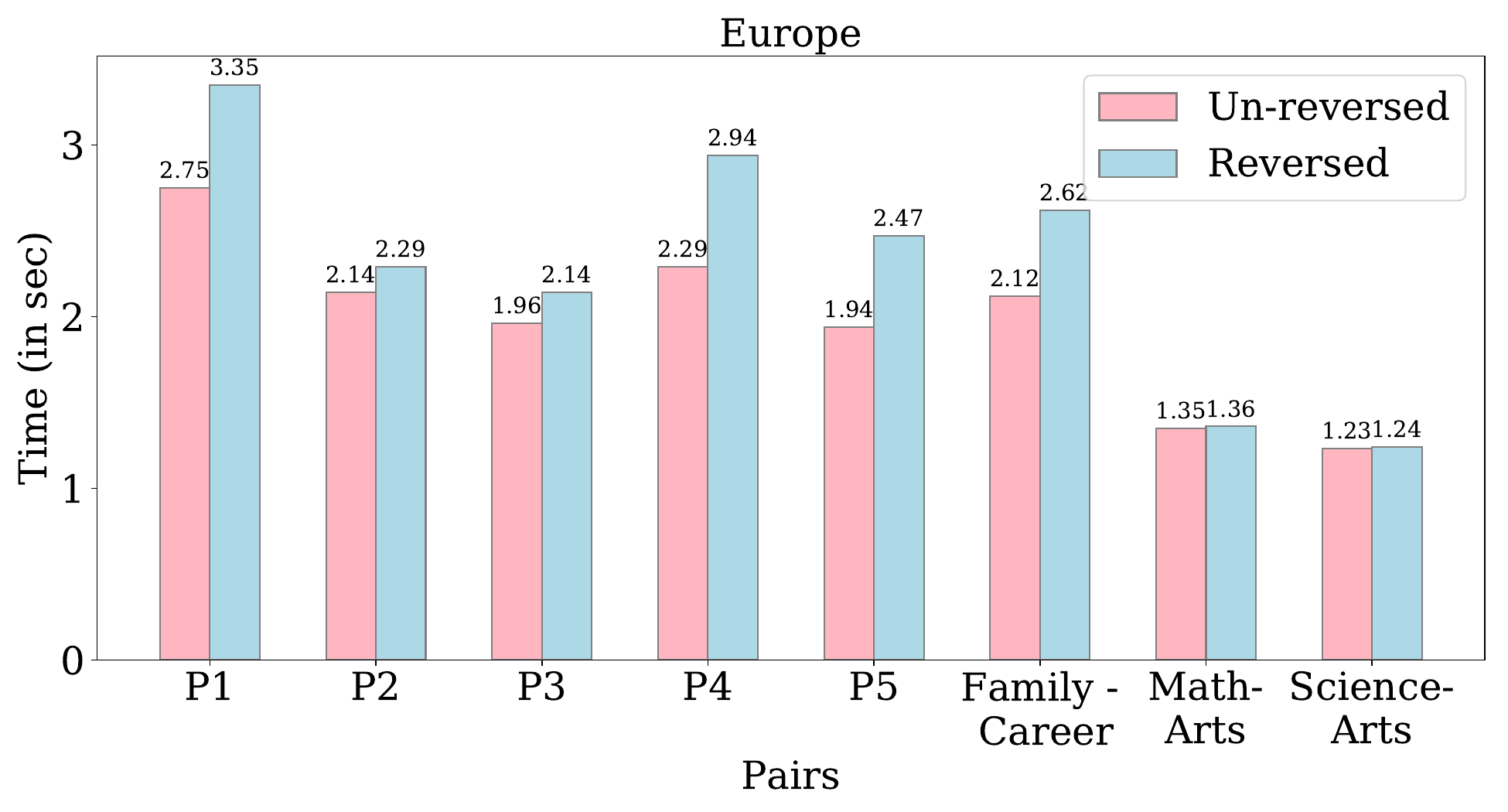}  
    \end{subfigure}  
    \begin{subfigure}{0.47\textwidth}
        \centering  
        \includegraphics[width=\textwidth]{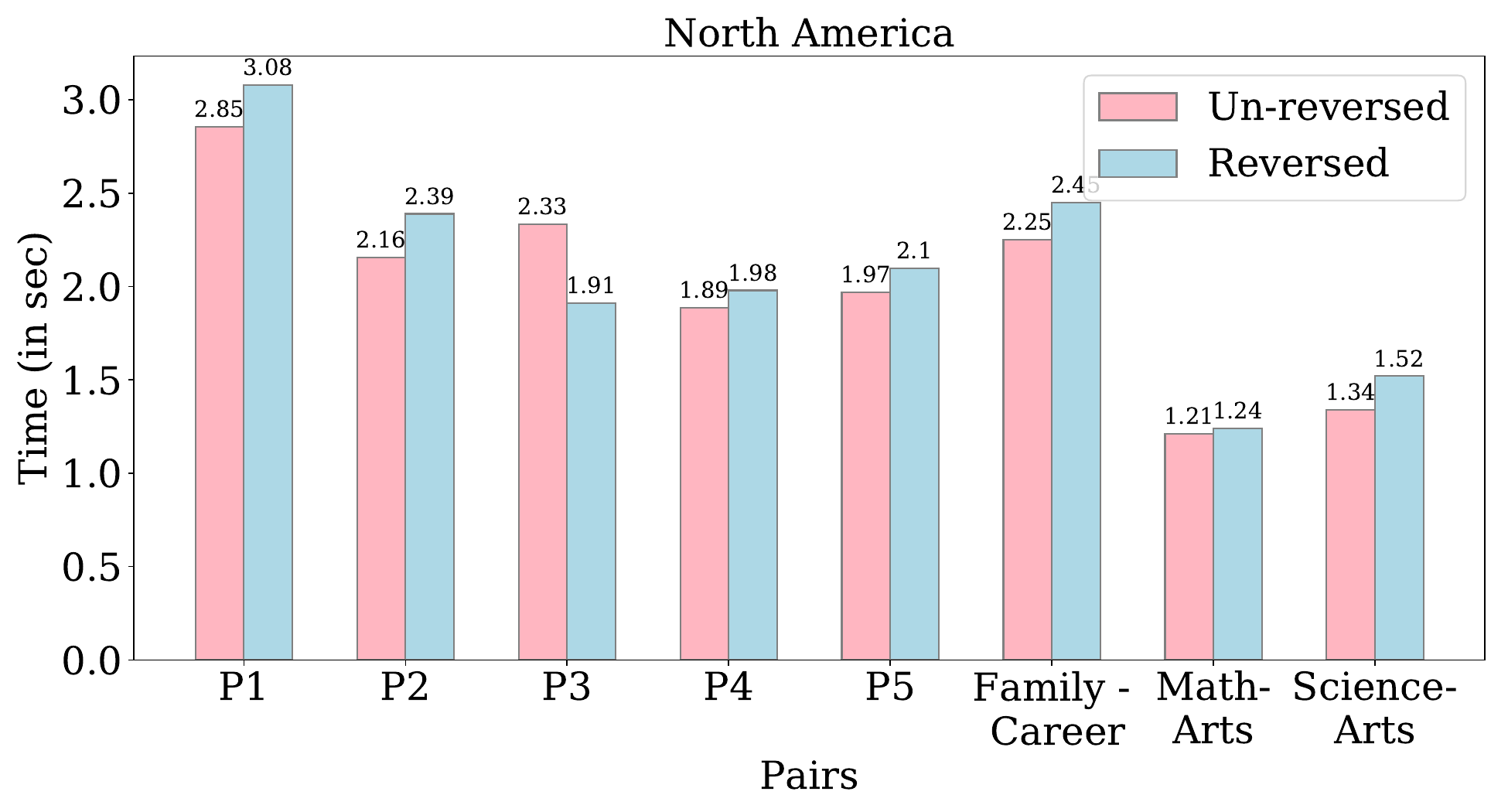}  
    \end{subfigure}  
     \vspace{-0.2cm}       
    \caption{\centering Human validation results across regions. `Unreversed' refers to bias dimensions with the same gender associations as our topic pairs, `Reversed' refers to bias dimensions with the opposite gender associations.}  
    \vskip -0.15in
    \label{fig:iat__}  
\end{figure*} 

\definecolor{pastelblue}{rgb}{0.678, 0.847, 0.902}

\subsubsection{Unigram/Bigram Analysis}

We find several topics that are common across regions. However, they may differ across cultures and may reveal varied perceptions of biases. Several topics also change associations to genders based on regions. For example, \textit{`fashion and lifestyle'} in Africa is associated with \textit{males}, however, it is associated with \textit{females} in Europe and North America. Several topics like \textit{`family and parenting'} are commonly associated with females across different regions while \textit{`politics'} is associated with males. To this end, we compute the top uni-grams and bi-grams for topic pairs that are common across regions in Appendix~\ref{sec:unibi}. 



\noindent
\subsubsection{Human Validation Results} 
Fig~\ref{fig:iat__} shows response times for top five topic pairs in each region for un-reversed and reversed scenarios. Larger time differences indicate more bias, suggesting that the pair could be a potential gender bias dimension for that region. If un-reversed time is lower, it suggests a stronger association of $T_f$ with the F group and $T_m$ with the M group. The family-career pair was also surveyed as a standard WEAT bias dimension. Please refer to Table~\ref{continent_topicpairs} for topic pair numbers (P1...P5) of each topic pair. 

\definecolor{PastelPurple}{RGB}{216, 191, 216}  
\begin{table*}[!ht]
\small
\centering
\begin{tabular}{l|p{9cm}|c|c}
\toprule
\centering \textbf{\textsc{Region}} &\centering  \textbf{\textsc{F-M topic pair}} & \centering  \textbf{\textsc{Reddit}} &\textbf{\textsc{UN General}}\\ 
& & & \textbf{\textsc{Debates}} \\
\midrule
 & Parenting and family relationships-Nollywood Actress and Movies & 0.500 & 0.979 \\ 
& Marriage and relationships - Sports and Football & -0.051 & 0.224 \\ 
Africa & Womens' lives and successes - Fashion and Lifestyle & 0.480 & 0.493  \\ 
& Music - Social Media & \colorbox{PastelPurple}{1.894} & \colorbox{PastelPurple}{1.721} \\ 
& Dating and relationships advice - Religious and Spiritual growth & 1.475 & 1.061 \\ \midrule   

 & Hotel royalty - Political leadership in India & 1.365 & \colorbox{PastelPurple}{1.768} \\ 
& Healthy eating habits for children - Sports and Soccer & 0.006 & -0.068 \\ 
Asia & Royal wedding plans - Social Media platforms for video sharing & 1.05 & 1.393 \\ 
& Royal wedding plans - Religious devotion and spirituality & 1.183 & 1.335 \\ 
& Marriage - Bollywood actors and films & \colorbox{PastelPurple}{1.543} & 0.918 \\ \midrule 

 & Education - Music & 1.261 & \colorbox{PastelPurple}{1.920} \\ 
& Comfortable hotels - Political decision and impact on society & 0.324 & 0.485 \\ 
Europe & Luxury sailing - UK Government Taxation policies & 1.232 & 1.558 \\
& Obituaries and Genealogy - Christian Theology and Practice & 0.001 & -0.405 \\ 
& Fashion and style - Christian theology and practice & \colorbox{PastelPurple}{1.730} & 1.028 \\ \midrule

 & Online Dating for Singles - Religion and Spirituality & \colorbox{PastelPurple}{1.728} & \colorbox{PastelPurple}{1.830} \\ 
& Fashion and Style - Reproductive Health & 1.723 & 1.095 \\ 
North America & Education and achievements - Reinsurance and capital markets & -0.148 & -0.364 \\ 
& Family dynamics and relationships - Nike shoes and fashion & 0.109 & 0.691 \\ 
& Reading and fiction - Cape Cod news & 0.251 & 0.506 \\ \midrule  

 & Family relationships - Religious beliefs and figures & 0.305 & 0.267 \\
& Woodworking plans and projects - Music record and Artists & 0.056 & -0.258 \\
Oceania & Weight loss and nutrition for women - Building and designing boats & 
0.336 & 0.582 \\ 
& Exercises for hormone development - Superheroes and their Universes & -0.05 &-0.07 \\ 
& Kids' furniture and decor - Building and designing boats & \colorbox{PastelPurple}{0.612} & \colorbox{PastelPurple}{0.524} \\ 
\bottomrule  
\end{tabular}
\caption{\label{reddit_topicpairs} 
Region-aware WEAT-based evaluation on Reddit and UNGDC. Highest scores are highlighted for each dataset across regions.
}
\vspace{-0.15in}
\end{table*}
As expected, the \textit{family--career} pair shows the highest bias across all three general IAT topic pairs. There are smaller differences among the other \textit{math--arts} and \textit{science--arts}. 
We also note that some pairs, such as {\it dating and relationships advice--religious and spiritual growth} (P5) for Africa, {\it hotel royalty--political leadership in India} (P1) for Asia,  {\it obituaries and geneology--Christian theology} (P4), {\it education--music} (P1), and {\it fashion and style--Christian theology} (P5) for Europe, and {\it online dating--religion and spirituality} (P1), {\it fashion and style--reproductive health} (P2) for North America have differences higher than those for {\it family--career} in the respective regions, indicating that the participants associated more biases on our uncovered bias dimensions than the existing one in WEAT.
These findings support our hypothesis that gender bias dimensions vary across regions and also bring preliminary evidence that the region-aware bias dimensions we uncover are in line with the human perception of bias in those regions. 
We also find that all the regions have biases that conform to our topic pairs gender association except P3: {\it education--reinsurance and capital markets} in North America, where the associated bias is negative. 
These findings confirm that topic pairs indeed differ across regions and that these differences must be taken into consideration when identifying and evaluating biases.

\section{WEAT-based Evaluation Using Region-aware Topic Pairs}

To measure biases in different data domains and regions, we extract region-aware topics using the GeoWAC dataset which spans Common Crawl separated by regions, and create a WEAT-style evaluation setup using these topics. 

\noindent \textbf{Data.} We consider two datasets: (i) Reddit data and (ii) UN General Debates \cite{doi:10.1177/2053168017712821}. The Reddit data consists of data from subreddits corresponding to specific regions: \texttt{r/asia, r/africa, r/europe, r/northamerica}, and \texttt{r/oceania}. We use the official Reddit API to extract data, consisting of 500 top posts\footnote{The Official Reddit API has rate limits, therefore 500 top posts from each subreddit ensures an equal number of examples for each region.} from each subreddit. The posts are pre-processed to remove URLs and signs, and each post contains at least 30 words. The UN General Debate Corpus (UNGDC) includes texts of General Debate statements from 1970 to 2016. These statements, similar to annual legislative state-of-the-union addresses, are delivered by leaders and senior officials to present their government's perspective on global issues. We filter the countries for each region and extract 500 data points per region, maintaining equal representation across region.\footnote{Oceania has limited available countries in UNGDC, hence the adherence to 500 data points for each region.}


\noindent \textbf{Method.} WEAT tests consist of keywords corresponding to each attribute and topic word sets like family-career and male-female terms. To create a similar setup, we utilize KeyBERT \cite{grootendorst2020keybert} to gather top topic representations corresponding to each topic extracted from GeoWAC. For male/female terms, we use the same representative words from WEAT. To further make it specific to a particular region, we employ GPT-4 \cite{openai2024gpt4} to generate common male/female names used in the regions and add them to the list. We provide the list of words in Table~\ref{tab:weat_list} of Appendix \ref{sec:weat_eval}. We use fastText \cite{bojanowski2016enriching}\footnote{We choose fastText because it allows to extract embeddings of words that are not present in the target text (as our topics are derived from GeoWAC).} embedding algorithm to generate embeddings of the lists and compute the distances between the topic words and male/female terms (like WEAT). 


\begin{figure}
\centering
  \includegraphics[width=7.5cm]{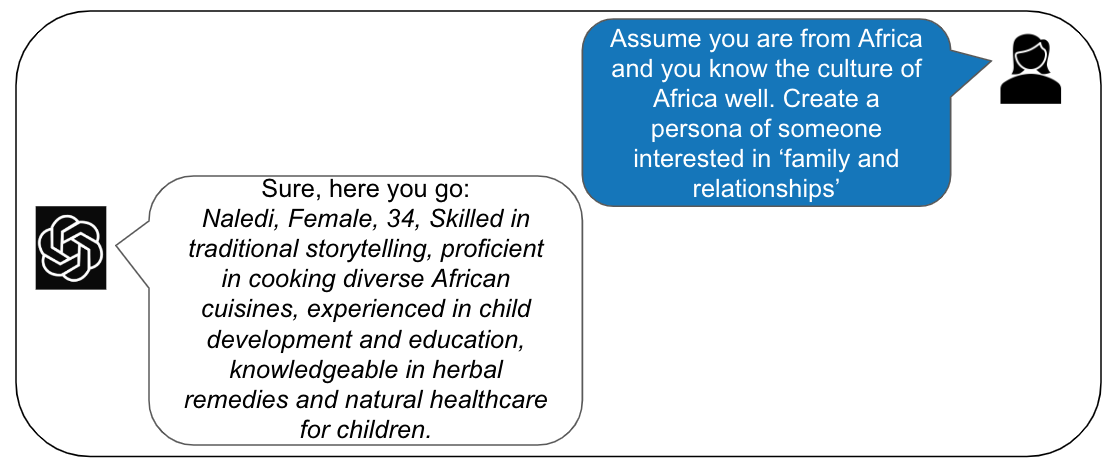}
\caption{Example Prompt for Persona Generation}
\label{fig:persona_eg}
\end{figure}

\begin{figure*}
    \centering
    \includegraphics[width=1\linewidth]{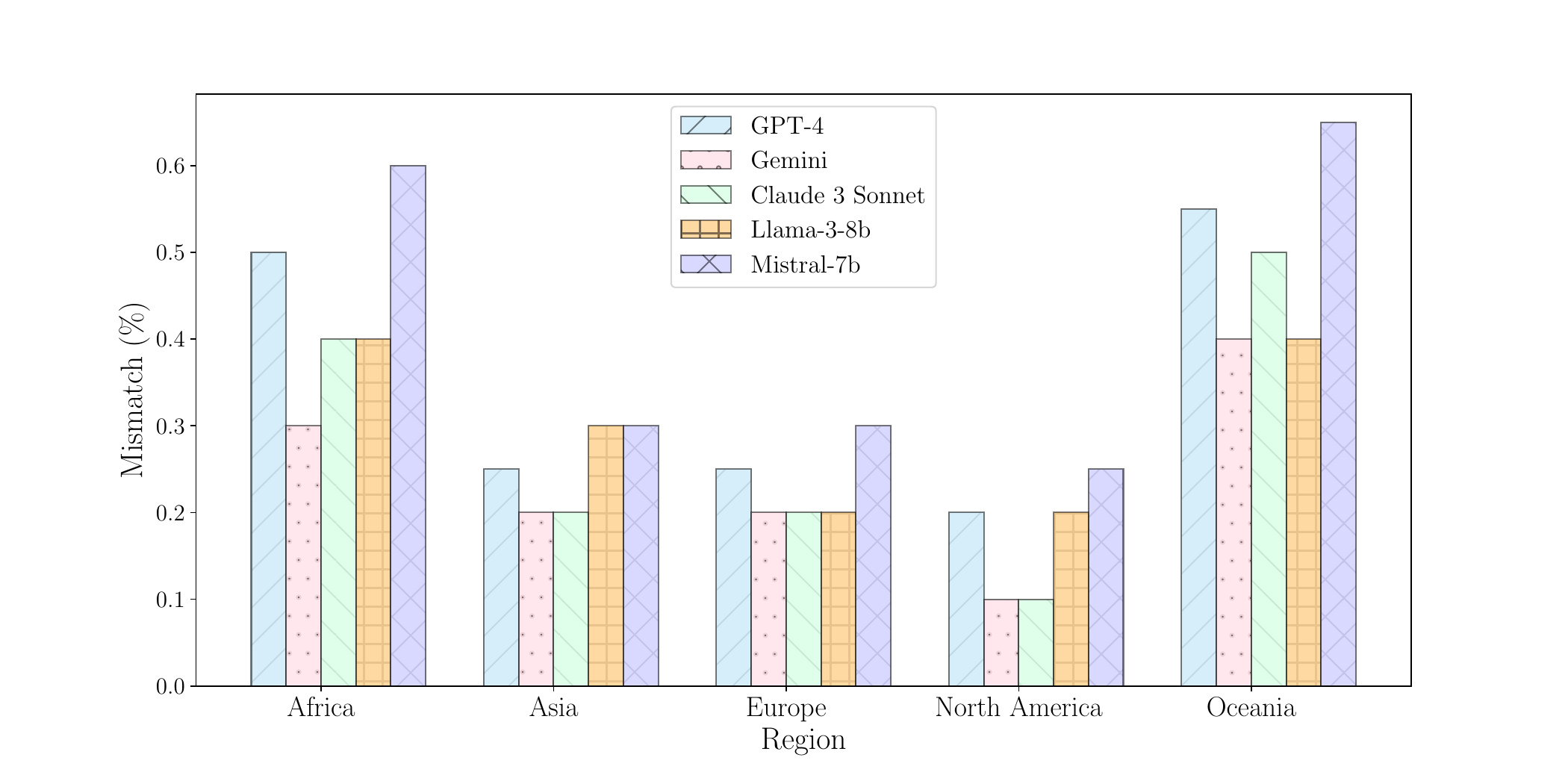}
    \caption{Bias Evaluation of LLM outputs using region-aware bias topic pairs through `persona generation'.}
    \label{fig:llms}
    \vspace{-0.15in}
\end{figure*}

\noindent  \textbf{Results.} 
In Table~\ref{reddit_topicpairs}, a high number of positive scores indicates the presence of a positive bias for our region-aware topic pairs. This means a presence of bias with the same gender association as our topic pairs, for example, if \textit{`music-social media'} is F-M topic pair in Africa according to our study, a positive score on the Reddit dataset means that bias is in the same direction. The few negative scores in the table indicate that these topic pairs do not conform to the same gender bias associations. However, a higher negative magnitude also shows the presence of bias, therefore, these topic pairs are still important. 

Additionally, magnitudes of many scores are high ($> 0.5$) which shows a high presence of bias (positive/negative) corresponding to the topic pairs. We highlight the top-scoring bias topic pairs for each region in Table~\ref{reddit_topicpairs}. High-bias topics vary for each region based on the dataset. For example, \textit{`music-social media'} has the highest bias in Africa for both datasets, however for Asia, we find that \textit{`marriage - Bollywood actors and films'} and \textit{`Hotel royalty - Political leadership in India'} are the topic pairs with the highest biases in Reddit and UN General Debates respectively, indicating that biased topic pairs may be domain-dependent.

Using our topic pairs in this WEAT-style evaluation setup provides an illustration of how our automatically curated region-aware bias dimensions can be used in designing a region-aware bias evaluation test.
It also shows the effectiveness of our region-aware bias topic pairs in capturing the dimensions that are likely to contain gender biases across regions.\footnote{Note that our topic pairs although extracted from GeoWAC are somewhat generalizable to other datasets like Reddit and UNGDC, we do not claim that these are best topic pairs achievable as topic pairs are also data dependent, but we can use our methodology to extract bias topic pairs that may exist in specific datasets.}
\section{Alignment of Region-Aware Bias Dimensions with LLM outputs}
To understand if LLMs generate similar biases as our region-aware bias topic pairs, we devise a persona generation task by LLMs. We prompt the LLM to output personas interested in different `topics' from the topic pairs that we extract. Fig~\ref{fig:persona_eg} shows an example of the prompt given to an LLM to generate personas.  We experiment with different LLMs:  \texttt{GPT-3.5} \cite{gpt35}, \texttt{GPT-4}, \texttt{Mistral-7b-Instruct} \cite{mistral7b}, \texttt{Claude-3 Sonnet},\footnote{https://claude.ai/} and \texttt{Gemini-Pro} \cite{geminiteam2024gemini}. Many studies use LLM-generated personas for multi-agent interactions in different settings in societies \cite{Park2023GenerativeAgents, zhou2024sotopia}. But, if an LLM generates biased personas, for example, a female persona takes care of children, and a male persona is strong and takes care of emergencies, this would lead to further biases in consequent tasks. Therefore, we employ persona generation to check for the presence of any biases in the personas created by LLMs. To measure biases, we find the number of matched LLM output persona genders to the genders of our topic pairs.
We average our results over seven runs.

\noindent {\bf Results.}
We plot the results of persona gender mismatched by LLMs in Fig~\ref{fig:llms}. The y-axis shows \% mismatch between the LLM generated persona gender and the gender of the topic in our topic pair. For example, a mismatch is when LLM outputs a persona with `female' for \textit{Politics} in Asia, which is a `male' topic according to our findings. Regions with high representation: North America, Europe and Asia have fewer mismatches, with North America having the lowest mismatch. Conversely, less represented regions like Africa and Oceania show higher mismatch rates. Among models, \texttt{Mistral-7b} (7B) has the highest mismatch rate while \texttt{Gemini-Pro} (50T) has the least, which may stem from varying model sizes.
Overall, all the models exhibit similar mismatch trends for both highly and less represented regions. Fewer mismatches in highly-represented regions show the importance of evaluation using region-specific topic pairs. Higher mismatches in underrepresented regions like Africa and Oceania suggest LLMs don't mimic these areas' biases, which can be beneficial. However, due to growing research on LLMs' cultural alignment, a more precise, region-specific bias evaluation metric becomes essential.


\section{Related Work}

IAT is one of the earliest and well-known method for measuring implicit social biases in humans \cite{greenwald1998measuring}. Inspired by the IAT, WEAT uses word embeddings to measure biases in text \cite{caliskan2017semantics}. Another extension of WEAT is the Sentence Embedding Association Test (SEAT), which measures biases at the sentence level \cite{may-etal-2019-measuring}. Additionally, various bias detection measures in NLP focus on post-training model predictions, such as gender swapping \cite{stanovsky-etal-2019-evaluating}. Moreover, there are specific gender bias evaluation test sets in tasks like coreference resolution \cite{rudinger-etal-2018-gender, zhao-etal-2018-gender, webster-etal-2018-mind} and sentiment analysis \cite{kiritchenko-mohammad-2018-examining}.

Several studies have emphasized the significance of considering cultural awareness in the study of social phenomena. The demographics of individuals can shape their worldviews and thoughts \cite{garimella-etal-2016-identifying}, potentially influencing their language preferences and biases in daily life. Notably, some studies have observed a bias towards Western nations in current LLMs \cite{dwivedi-etal-2023-eticor}. Recent research has focused on cross-cultural aspects of LLMs, including aligning them with human values from different cultures \cite{glaese2022improving, sun2023principle} and exploring them as personas representing diverse cultures \cite{gupta2024selfassessment}. To the best of our knowledge, no previous work has proposed a data-dependent approach to extract region-aware bias topics. Given the known biases in LLMs, a region-specific metric could greatly lead to an accurate evaluation of biases. This research holds significant importance in addressing cross-cultural biases effectively.

\section{Conclusion}

In this paper, we proposed a bottom-up approach using data to identify region-aware topic pairs that capture gender biases across different regions. Our human evaluation results demonstrated the validity of our proposed region-aware dimensions. 

We employed a region-aware WEAT-based evaluation setup to assess biases in two additional datasets: Reddit and the UN General Debate Corpus. The presence of region-specific biases in these datasets underscores the importance of a region-aware bias evaluation metric. Additionally, when examining LLM outputs against the gender associations in our region-aware bias topic pairs, we found that biases align closely for three highly represented regions: North America, Europe, and Asia. This emphasizes the value of region-aware topic pairs in bias evaluation of LLMs. Conversely, biases do not align well for Africa and Oceania, indicating that LLMs do not adopt these regions' specific biases--a potential benefit. Yet, it also highlights the 'cultural alignment' issue in LLMs. More research on the cultural alignment of LLMs underlines the need to consider region-specific bias topic pairs for all regions in future studies.

Future work includes incorporating testing different model/dataset combinations and topic-pair dependency on data. We aim to study biases in different languages and explore region-aware bias mitigation techniques.


\section{Limitations}
We utilized the GeoWAC corpus as our sole data source for extracting topic pairs from various regions. However, we acknowledge the importance of incorporating additional datasets in our future work. Additionally, our WEAT-based evaluation was conducted on relatively smaller datasets. So, we intend to conduct further analysis on a larger dataset to ensure a comprehensive evaluation based on WEAT.

Our study did not account for different languages due to the diverse linguistic landscape of the regions (continents) included in our study. However, the significance of conducting a more detailed analysis to examine variations among different countries would be interesting.

Unfortunately, we encountered difficulties in finding participants from Oceania for human validation. Moving forward, we plan to include insights and findings from Oceania and incorporate a larger population to ensure a more comprehensive human validation.

\section{Ethical Considerations}
When developing our region-aware topic pairs, it is essential to consider the ethical implications. Since we utilize a much broader aspect of culture, i.e. continents to distinguish among cultures, the region-aware topic pairs we extract may not translate to cultures of communities that are not well-represented in models. Hence, it is important that we utilize topic pairs carefully. 

It has been found that AI models often tend to output responses that are Western, educated, industrialized, rich, and democratic \cite{henrich2010weirdest}. In our experiments, we see LLMs also generate biases having the highest alignment with the West. Therefore, LLM experiments also need to be utilized carefully. 

Our Reddit data for the region-aware evaluation metric may contain offensive content. However, we have anonymized the data (removed the usernames). 

\bibliography{cross_cultural, anthology1}

\appendix

\begin{table}[ht]
\small
\centering
\begin{tabular}{p{2cm}|p{1.3cm}|p{1.7cm}}
\toprule
\textbf{\textsc{Region}} & \textbf{\textsc{Country}} & \textbf{\textsc{\#Examples}} \\ \midrule 
  & Nigeria & 3,153,761 \\ 
 Africa & Mali & 660,916 \\
 & Gabon & 645,769 \\ \midrule 
 
   & India & 12,327,494 \\ 
 Asia  & Singapore & 6,130,047 \\ 
  & Philippines & 3,166,971 \\ \midrule 

    & Ireland & 8,689,752 \\ 
 Europe  & United Kingdom & 7,044,434 \\ 
  & Spain & 465,780\\ \midrule 

  & Canada & 7.965,736 \\ 
 North America & United States & 8,521,094 \\ 
  & Bermuda & 244,500 \\ \midrule

  & New Zealand & 94,476 \\ 
 Oceania  & Palau & 486,437 \\ 
  & Vanuatu & 165,355 \\ \bottomrule 
  
\end{tabular}
\caption{\label{geowac_details}
Region-specific details in GeoWAC
}
\end{table}


\section{GeoWAC dataset details}
\label{geowac_data}
Table~\ref{geowac_details} contain the total number of examples per country in a region. We consider the top three countries with the highest number of examples per region. 

\section{F-M Dataset statistics}
\label{fm_appendix}
Table~\ref{Dataset_Cont} displays the total number of examples from female and male groups per region for the region-specific F-M dataset. 

\begin{table}[ht]
\small
\centering
\begin{tabular}
{p{2cm}|p{1.3cm}|p{1.3cm}|p{1.3cm}}
\toprule
\textbf{\textsc{Region}} & \textbf{\textsc{Total}} & \textbf{\textsc{\#Female}} & \textbf{\textsc{\#Male}} \\ \midrule 
 Africa & 57895 & 20153 & 37742 \\ 
 Asia & 56877 & 21400 & 35477 \\ 
 Europe & 59121 & 21049 & 38072 \\ 
 North America & 70665 &27627 & 43038\\  
 Oceania & 62101  &25951 & 36150\\ \bottomrule 
\end{tabular}
\caption{\label{Dataset_Cont}
F-M dataset statistics for regions (Total refers to the total number of examples in each region, therefore, total = \#female + \#male)
}
\end{table}

\begin{table*}
\scriptsize
\begin{tabular}{lcccp{2cm}}  
\toprule  
\textbf{\textsc{Target words - Attribute words}} & \textbf{\textsc{Region}} & \textbf{\textsc{Region-specific p-value}} & \textbf{\textsc{Region-specific WEAT score}} & \textbf{\textsc{Original WEAT score, p-value}} \\  
\midrule  
 & Africa & 0.016 & 1.798 & 1.81, 0.001 \\  
 & Asia & 0.007 & 1.508 & \\  
 Male names vs Female names & North America & 0.04 & 1.885 & \\  
 - career vs family & Europe & $6 \cdot 10^{-4}$ & 1.610 &  \\  
 & Oceania & 0.03 & 1.727 & \\  
\midrule  
  & Africa & 0.003 & 1.429 & 1.06, 0.018 \\  
 & Asia & 0.045 & 1.187 & \\  
 Math vs Arts & North America & 0.007 & 0.703 & \\  
 - Male vs Female terms & Europe & 0.005 & 0.334 & \\  
 & Oceania & 0.03 & 1.158 & \\  
\midrule  
  & Africa & 0.048 & 1.247 & 1.24, 0.01 \\  
 & Asia & 0.004 & 0.330 & \\  
 Science vs Arts & North America & $1 \cdot 10^{-5}$ & 0.036 & \\  
 - Male vs Female terms & Europe &  $1 \cdot 10^{-7}$ & \colorbox{PastelPurple}{-0.655}  & \\  
 & Oceania & $2 \cdot 10^{-4}$ & 0.725  & \\  
\midrule  

 & Africa & $3 \cdot 10^{-5}$ & 0.855 & 1.21, 0.01 \\  
Young people names & Asia & $4 \cdot 10^{-4}$  & 0.917 &  \\  
 vs old people names & North America & 0.032 & 1.325 & \\  
- pleasant vs unpleasant & Europe & 0.009 & 0.917 & \\  
  & Oceania & 0.014 & 0.947 & \\  
\midrule  

   & Africa & $1\cdot 10^{-5}$ & 0.008 & 1.28, 0.001 \\  
 European American names & Asia & $1 \cdot 10^{-6}$ & \colorbox{PastelPurple}{-0.453}  & \\  
 vs African American names & North America & 0.009 & 1.29 & \\  
- pleasant vs unpleasant & Europe & 0.001 & 0.617 & \\  
 & Oceania & $1\cdot 10^{-4}$  & 0.492  & \\  
\midrule  

  & Africa & 0.03 & 1.443 & 1.53, $< 10^{-7}$ \\  
 & Asia & 0.009 & 1.001 & \\  
Instruments vs Weapons & North America & 0.01 & 1.202 & \\  
 - pleasant vs unpleasant & Europe & 0.02 & 1.21 & \\  
 & Oceania & 0.001 & 0.951 & \\  
\midrule  

 & Africa & 0.002 & 0.312 & 1.5, $< 10^{-7}$ \\  
 & Asia & 0.009 & 0.869 & \\  
 Flowers vs Insects  & North America & 0.003 & 0.382 & \\  
- pleasant vs unpleasant & Europe & 0.001 & 0.332 & \\  
 & Oceania & 0.009 & 0.660 & \\  
\midrule  

  & Africa & 0.008 & 0.835 & 1.38, 0.01 \\  
 & Asia & 0.02 & 1.201 & \\  
 Mental disease vs Physical disease & North America & 0.008 & 0.692 & \\  
 - temporary vs permanent & Europe & 0.04 & 1.382 & \\ 
 & Oceania & 0.009 & 1.620 & \\ 
\bottomrule  
\end{tabular}
\caption{\label{weat_cont_all}Region-wise WEAT scores and p-values across all dimensions specific in WEAT using word2vec. Negative scores are highlighted. We compare our region specific scores and p-values with the scores and p-values of the Original paper by \cite{caliskan2017semantics}}
\end{table*}

\section{Cultural differences in biases using WEAT}
Table~\ref{weat_cont_all} shows the WEAT scores for all WEAT dimensions defined in \cite{caliskan2017semantics}. We find that scores and p-values differ across regions for different dimensions. High bias dimensions differ across regions, hence it is important to consider region-specific topic pairs.

\section{Region-wise topic lists in GeoWAC}
\label{topiclist_appendix}

Table~\ref{continent_toptopics_all} displays a comprehensive list of topics for female and male groups across all regions. 
\begin{table*}[ht]  
\scriptsize  
\centering  
\begin{tabular}{|p{3cm}|p{4cm}|p{4cm}|}  
\hline  
\textbf{\textsc{Region}} & \textbf{\textsc{Female}} & \textbf{\textsc{Male}} \\ \hline  
  
Africa & Credit cards and finances, Royalty and Media, Trading strategies and market analysis, Dating and relationships guides, Parenting and family relationships, Fashionable Ankara Styles, women's lives and successes, online dating & Fashion and Lifestyle, Male enhancement and sexual health, Nollywood actresses and movies, Nigerian politics and government, Essay writing and research, Medical care for children and adults, Journalism and Media Conference, Music industry news and releases, Football league standing and player performances, Academic success and secondary school education, Religious inspiration and spiritual growth, Economic diversification and Socio-economic development \\ \hline  
  
Asia & Hobbies and Interests, Healthy eating habits for children, Social media platforms, Royal wedding plans, Online Dating and Chatting, Adult Services, Gift ideas for Valentine's Day & DC comic characters, Mobile Application, Philippine Politics and Government, Sports and Soccer, Career, Bike enthusiasts, Artists and their work, Youth Soccer Teams, Career in film industry, Political leadership in India, Bollywood actors and films, Religious devotion and spirituality, Phone accessories \\ \hline  
  
Europe & Pets and animal care, Fashion and Style, Education, Obituaries and Genealogy, Luxury sailing, Traveling, Energy and climate change, Family and relationships, Pension and costs, Tech and business operations, Dating, Comfortable hotels, Government transportation policies  & Political developments in Northern Ireland, Christian Theology and Practice, Crime and murder investigation, EU Referendum and Ministerial Positions, Criminal Justice System, Israeli politics and International relations, Cancer and medications, UK Government Taxation policies, Art Exhibitions, Political decision and impact on society, Music Gendres and artists, Medical specialties and university training, Political discourse and parliamentary debates  \\ \hline  
  
North America & Pets, Cooking: culinary delights and chef recipes, Fashion and style, Family dynamics and relationships, Reading and fiction, Scheduling and dates, Life and legacy of Adolf Hitler, Gender roles and inequality, Education and achievements, Online dating for singles, Luxury handbags, Footwear and Apparel brands, Essay writing and literature & Civil War and history, Middle East conflict and political tensions, Movies and filmmaking, Political leadership and party dynamics in Bermuda, Rock Music and songwriting, Wartime aviation adventures, Religion and Spirituality, Reproductive health, Reinsurance and Capital markets, Nike shoes and fashion, Cape Cod news, NHL players \\ \hline  
  
Oceania & Cooking and culinary delights, Romance, Weight loss and nutrition for women, Water travel experience, Woodworking plans and projects, Time management and productivity, Inspiring stories and books for alleges, Sexual violence and abuse, Car insurance, Exercises for hormone development, kid's furniture and decor & Harry Potter adventures, Art and Photography, Superheroes and their Universes, Music recording and Artists, Football in Vanuatu, Pet care and veterinary services, Building and designing boats, Religious beliefs and figures, Fashion, Classic movie stars, Men's hairstyle and fashion, Male sexual health and supplements  \\ \hline  
  
\end{tabular}  
\caption{\footnotesize Region-wise topics for female and male.}  
\label{continent_toptopics_all} 
\vspace{-0.15cm}
\end{table*}

\begin{table}[ht]  
\centering  
\scriptsize
\begin{tabular}{p{2cm}|p{2cm}|p{2cm}}  
\toprule  
\textsc{\textbf{Region}}      & \textsc{\textbf{Female topics}  }                        & \textsc{\textbf{Male topics}}                                 \\  
\midrule  
Africa       & Credit card-based financial services     & Fashion - footwear and celebrities          \\  
              & Royalty and femininity                   & Male enhancement and sexual health          \\  
              & Financial trading                        & Nollywood                                   \\  
              & Dating guides                            & Nigerian politics                           \\  
              & Motherhood and parenting                 & Academic writing                            \\  
\midrule
Asia         & Hobbies                                 & Superhero comic books                       \\  
              & Food and nutrition                       & Mobile applications                         \\  
              & Social media platforms and content creation & Philippines politics and people          \\  
              & Royal weddings                           & Sports                                      \\  
              & Online social interaction and dating     & Career                                      \\  
              \midrule

Europe       & Pets                                    & Irish politics                              \\  
              & Fashion                                  & Christianity                               \\  
              & Education                                & Law enforcement and crime                   \\  
              & Deaths and funerals                      & EU and Brexit                               \\  
              & Luxury yachting and sailing              & Criminal justice system                     \\  
              \midrule

North America& Pets                                    & Civil War Military                          \\  
              & Cooking and Food                         & Middle Eastern politics and conflicts        \\  
              & Fashion                                  & Movies and direction                        \\  
              & Family and relationships                 & Bermuda politics                            \\  
              & Reading novels                           & Rock music                                  \\ 
              \midrule

Oceania      & Food and eating habits                   & Harry Potter                                \\  
              & Romance and emotions                     & Artistic expressions                        \\  
              & Weight loss and nutrition                & Superheroes of Marvel and DC                \\  
              & Boat and sailing experience              & Albums, songs and artists                   \\  
              & Woodworking and carpentry                & Vanuatu Football                            \\  
\bottomrule  
\end{tabular}
\caption{\label{gpt_topics}
Topic labels by \texttt{gpt-4}, see Table~\ref{continent_toptopics} for comparison with \texttt{Llama2} topic labels}
\end{table}

\begin{figure*}
\centering
  \includegraphics[width=15cm]{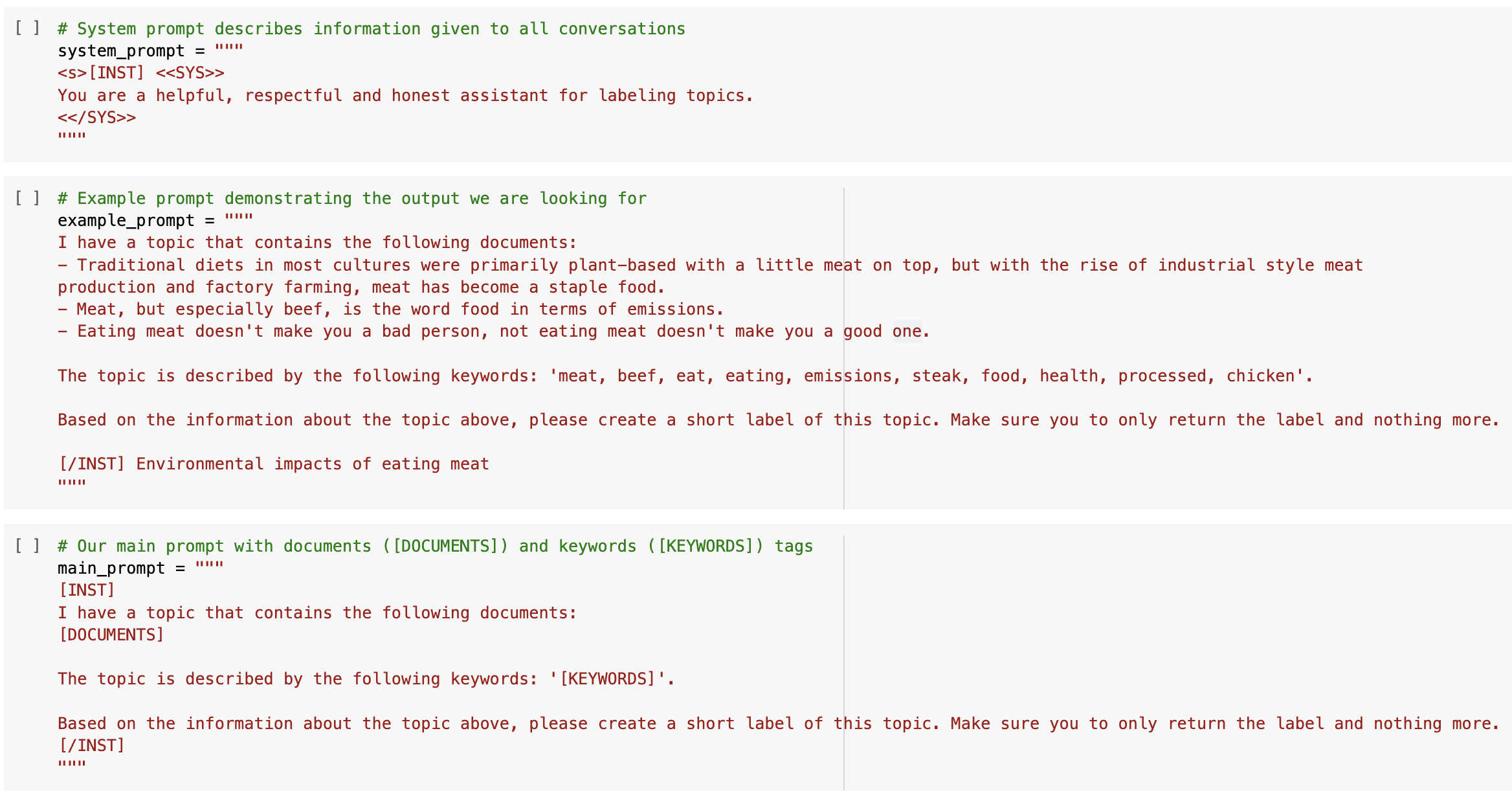}
\caption{\texttt{Llama2} prompt}
\label{fig:prompt}
\end{figure*}

\section{Unigram/Bigram Analysis}
\label{sec:unibi}
\begin{table*}[ht]  
\centering
\scriptsize
\begin{tabularx}{\textwidth}{c|p{1.5cm}|p{4.1cm}|X}  
\toprule  
\textbf{\textsc{Topic}} & \textbf{\textsc{Region}} & \textbf{\textsc{Unigrams}} & \textbf{\textsc{Bigrams}} \\  
\midrule  
& Africa (male) & march, outlet, air, max,  tods, man, said, pas, cher, people & air max, pas cher, princess j, roshe run, nike air, tods outlet, j march, roger vivier, posts email, notify new\\  
\cmidrule{2-4}  
\textbf{Fashion and lifestyle} & Europe (female) & one, women, fashion, like, new, look, make, hair, girl, dress & oakley sunglasses, louis vuitton, red carpet, new york, fashion model, engagement rings, per cent, year old, christian louboutin, diamond ring \\  
\cmidrule{2-4}  
& North America (female) & one, love, like, little, new, made, time, get, make, women & s cooper, cooper main, t shirt, new york, little girl, men women, look good, main store, years ago, check out\\  
\midrule 

& Africa (female) & music, song, album, new, video, single, one, singer, also, songs & music industry, hip hop, record label, single titled, new single, chris brown, tiwa savage, ice prince, kanye west, niegrian music\\  
\cmidrule{2-4}  
\textbf{Music} & Europe (male) & man, single, stage, years, world, many, metal, guitar, solo, irish & year shelfmark, black metal, time exercise, musical content, dundee repertory, singer songwriter, edinburgh year, zumba days, male vocalists, millions men \\  
\cmidrule{2-4}  
& Oceania (male) & music, album, new, songs, band, first, time, jazz, released, rock &  new york, elizabth ii, debut album, years later, big band, rock roll, first time, studio album, los angeles, solo artist \\  
\midrule 

\end{tabularx}  
\caption{\label{tab:common_diff}
Common topics with different gender associations across regions}
\end{table*}  

Table~\ref{tab:common_diff} shows the unigrams and bigrams of common topics with different gender associations. We find that `fashion' is highly associated with shoes when it is a male topic in Africa, whereas in Europe and North America, it is mostly associated with accessories like sunglasses, rings, etc. This shows the typical association of women with jewelry and men with shoes \cite{jewel, shoes}. In the case of `Music', we see that unigrams and bigrams pertaining to Africa contain words related to hip-hop music and artists. For Europe, we find location references and metal music. And finally, Oceania shows references of jazz and rock. We do not find any obvious gender associations in the analysis of the music topic. 
Table~\ref{tab:common_same} provides a unigram/bigram analysis of topics that are commonly associated with a specific gender across regions. For \textit{parenting and family relationships}, Africa has mentions of children, while Asia and Oceania contain mentions of family events, etc. In North America, we mostly find text about maintaining health in families. 
For \textit{religion and spirituality}, the unigrams/bigrams are mostly about Jesus and Christianity across regions. For \textit{politics}, we find mentions of specific regions, as expected. \textit{Education} topic is more about being successful in Europe, where it is about degrees in North America. Finally, `social media' trends are mostly similar. Overall for topics with same gender associations across regions, do not have stark differences.

\begin{table*}[ht]  
\centering
\scriptsize
\begin{tabularx}{\textwidth}{p{1.8cm}|p{2.2cm}|p{4.1cm}|X}  
\toprule  
\textbf{\textsc{Topic}} & \textbf{\textsc{Region}} & \textbf{\textsc{Unigrams}} & \textbf{\textsc{Bigrams}} \\  
\midrule  
& Africa (female) & child, registration, form, information, sent, women, foster, best, catholic, women & registration form, form information, child assigned, surgery doctors, new catholic, catholic women, contemporary challenge, best everything, foster short, doctors clinic\\  
\cmidrule{2-4}  
\textbf{Parenting and family relationships} & Asia (female) & year, old, weekly, fortnightly, clicking, create, alert, state, 1, terms & year old, weekly fortnightly, create alert, stated agree, conditions acknowledge, finals appearances, together playing, dial guarded, came work, outlet jackets \\  
\cmidrule{2-4}  
& North America (female) & women, healthday, loss, three, worked, closely, together, she, elegant, dignified  & three women, women worked, closely together, elegant dignified, very pleasant, soft spoken, women men, healthday reporter, tuesday march, participate more \\  
\cmidrule{2-4}  
& Oceania (female) & laurel, school, moved, one, day, royal, wedding, house, sister, hopefully & moved one, royal wedding, laurel school, 1 california, weeks dad, high school, one hopefully, nobody knew, sister means, fu school \\  
\midrule 

& Africa (male) & god, man, church, one, life, people, jesus, us, lord, christ,  & short description, jesus christ, man god, holy spirit, god said, thank god, bible says, catholic church, today god, every man\\  
\cmidrule{2-4}  
& Asia (male) & life, jesus, us, church, one, man ,lord, said, father, christ & holu spirit, jesus christ, pope francis, brothers sisters, son god, men women, holy father, opus dei, eternal life, paul ii\\  
\cmidrule{2-4}  
\textbf{Religion and Spirituality} & Europe (male) & god, one, jesus, church, life, people, father, man , said, christ & jesus christ, son man, catholic church, holy spirit, men women, said him, holy father, john paul, jesus said, word god  \\  
\cmidrule{2-4}  
& North America (male) & god, jesus, one, man, us, life, would,  christ, lord, people & recognizable cheering, section league, jesus christ, exact synonyms, past years, god said, years before, thanks mostly, mostly steph, father dell \\  
\cmidrule{2-4}  
& Oceania (male) & also, said, best, love, new, come, good, like, men, made &  god said, jesus christ, holy spirit, lord krishna, temple god, father devil, eternal life, son god, son man, god father \\  
\midrule

 & Asia (male) & said, one, India, time, people, minister, government, years, state, police, court & indian congress, government plans, modi ministry, human rights, foreign politics, armed forces, international warfare, foreign ministry, middle east,  united nations \\  
\cmidrule{2-4}  
 \textbf{Politics} & Europe (male) & government, said, minister, people, international, country, one, foreign, president, state & make statement, prime minister, human rights, armed forces, secretary state, middle east, united nations, hon friend, foreign secretary, united states \\  
\midrule

 &  Europe (female) & school, primary, teacher, founder, CEO, judgment, group, named, ranking, prestigious & as founder, founder CEO, judgment group, named fortune, ranking prestigious, world scientist, scientist women, students comprehend, program support, support students \\  
\cmidrule{2-4}  
 \textbf{Education} & North America (female) & bachelor, years, student, leader, degree, animal, veterinary, music, taught, communication & bachelors degree, animal veterinary, bachelor music, alison taught, privately years, students ranging, development programmes, including leader, art communication, recent years \\ 
\midrule 
 &  Africa (male) & onigbinder, aura, pictures, first, gained, popularity, match, beaut, designed, music & aura pictures, gained popularity, match beaut, designed wonder, attending music, music festival, schomburg library, Instagram account, sugar coating, schedule tomorrow \\  
\cmidrule{2-4}  
 \textbf{Social Media} & Asia (male) & time, later, latest, tracks, speedy, Zulfiqar, nasty, children, tweeted, guys & gets later, latest tracks, speedy zulfiqar, children pti, pti tweeted, taking long, long time, hosted pageant, time vincent, love fleeting \\ 

\bottomrule 

\end{tabularx}  
\caption{\label{tab:common_same}
Common topics with same-gender associations across regions}
\end{table*}

\section{WEAT-based evaluation setup details}
\label{sec:weat_eval}
For male/female terms, we use the same representative words from WEAT: \textit{brother, father, uncle, grandfather, son, he, his, him, man, boy, male} for male and \textit{sister, mother, aunt, grandmother, daughter, she, hers, her, woman, girl, female for female}.
We also utilize GPT-4 to output the ten most common male/female names specific to each region. We provide the lists of word belonging to 
each topic in Table~\ref{tab:weat_list}. 

\begin{table*}
\scriptsize
\centering
\begin{tabular}{lp{13cm}}  
\toprule  
\textsc{\textbf{Region}} & \textsc{\textbf{Topics: Word lists}} \\  
\midrule  
 & \textbf{Nollywood Actress and Movies}: nollywood, actress, actors, drama, celebrity, movie, acting, movies, producer, tv \\ 
& \textbf{Parenting and family relationships}: mother, mom, mothers, mum, moms, parent, her, child, momodu, parents \\
&\textbf{Sports and Football}: players, sports, fifa, team, player, football, mourinho, scored, league, champions\\
&\textbf{Marriage and relationships}: wives, marriage, husbands, marriages, married, wife, relationships, husband, marry, relationship\\
&\textbf{Fashion and lifestyle}: cher, nike, max, air, looked, face, love, tods, soldes, scarpe\\
AFRICA &\textbf{Womens lives and successes}: women, ladies, woman, female, girls, men, gender, ones, employees, male\\
&\textbf{Social Media}: instagram, facebook, social, twitter, tweet, snapchat, tweets, tweeted, hashtag, followers\\
&\textbf{Music}: song, songs, album, hits, music, released, rap, singer, tracks, rapper\\
&\textbf{Religious and Spiritual Growth}: god, almighty, bible, christ, faith, believers, christian, jesus, prayer, religion\\
&\textbf{Dating and relationships advice}: dating, women, relationships, ladies, sites, singles, online, single, escorts, websites \\
&\textbf{Male terms}: male, man, boy, brother, he, him, his, son, Kwame, Mandela, Moyo, Jelani, Tariq, Keita, Obi, Simba, Ayo, Kofi, Jabari, Tunde, Mekonnen, Anwar, Chukwuemeka \\
&\textbf{Female terms}: sister, mother, aunt, grandmother, daughter, she, hers, her, Aisha, Zahara, Nia, Sade, Amara, Chinelo, Layla, Ayana, Nala, Zuri, Imani, Lola, Kamaria, Nyala, Kaya \\
\midrule
 & \textbf{Political Leardership in India}: modi, political, said, bjp, told, says, leader, congress, minister, public\\
&\textbf{Hotel Royalty}: visited, places, stayed, hotels, adventure, pictures, favourite, guest, hiking, hemingway\\
&\textbf{Sports and Soccer}: sports, team, basketball, players, nba, league, championship, coach, rebounds, finals\\
&\textbf{Healthy eating habits for children}: food, foods, eating, meals, nutrition, cuisine, diet, dishes, cooking, eat\\
&\textbf{Social Media platforms for video sharing}: instagram, video, videos, twitter, tweet, facebook, gifs, vlog, youtube, followers\\
ASIA &\textbf{Royal wedding plans}: meghan, duchess, engagement, england, royal, royalty, prince, kate, london, married\\
&\textbf{Religious devotion and spirituality}: god, bible, holy, faith, prayer, believe, christian, blessed, christ, spiritual\\
&\textbf{Royal wedding plans}: meghan, duchess, engagement, england, royal, royalty, prince, kate, london, married\\
&\textbf{Bollywood actors and films}: bollywood, bachchan, kapoor, actors, acting, kareena, actor, film, shahrukh, hindi\\
&\textbf{Marriage}: married, marriage, marriages, couple, couples, wife, marry, wedding, husband, divorced\\
&\textbf{Male terms}: male, man, boy, brother, he, him, his, son, Hiroshi, Ravi, Kazuki, Jin, Satoshi, Rohan, Haruki, Dai, Akira, Yuan\\
&\textbf{Female terms}: sister, mother, aunt, grandmother, daughter, she, hers, her, Sakura, Mei, Aiko, Yuna, Lina, Ji-hye, Mika, Nami, Anika, Rina\\
\midrule
& \textbf{Music}: music, songs, vocalists, album, albums, singing, vocals, singles, rock, song\\
&\textbf{Education}: school, schools, classroom, students, education, educational, pupils, boys, academy, college\\
&\textbf{Political decisions and impact on society}: government, public, minister, said, hon, people, first, the, column, committee\\
&\textbf{Comfortable hotels}: guests, staying, rooms, friendly, welcoming, stayed, hotel, beds, stay, comfortable\\
&\textbf{UK Government Taxation Policies}: corbyn, taxation, fiscal, tax, taxes, exchequer, labour, governments, government, deficit\\
EUROPE &\textbf{Luxury Sailing}: yachts, yacht, boat, sailing, sails, cruising, sail, berths, cruiser, cabin\\
&\textbf{Christian Theology and Practice}: god, bible, christ, jesus, faith, christian, religious, religion, holy, gave\\
&\textbf{Obituaries and Genealogy}: died, edward, relatives, anne, lived, elizabeth, funeral, irish, mrs, galway\\
&\textbf{Christian Theology and Practice}: god, bible, christ, jesus, faith, christian, religious, religion, holy, gave\\
&\textbf{Fashion and style}: fashion, shoes, style, clothes, clothing, shoe, wear, nike, dress, stylish\\
&\textbf{Male terms}: male, man, boy, brother, he, him, his, son, Lukas, Matteo, Sebastian, Alexander, Gabriel, Nikolai, Maximilian, Leonardo, Daniel, Adrian\\
&\textbf{Female terms}: sister, mother, aunt, grandmother, daughter, she, hers, her, Emma, Sophia, Olivia, Isabella, Ava, Mia, Charlotte, Amelia, Lily, Emily\\

\midrule
 & \textbf{Religion and Spirituality}: god, christ, jesus, bible, christian, holy, christians, scripture, faith, heaven\\
&\textbf{Online Dating for Singles}: dating, singles, hookup, single, relationships, dates, flirting, personals, date, mingle\\
&\textbf{Reproductive Health}: download, available, pdf, online, edition, manual, free, reprint, kindle, file\\
&\textbf{Fashion and style}: fashion, dresses, dress, wardrobe, clothes, clothing, style, outfit, vintage, wear\\
&\textbf{Reinsurance and capital markets}: reinsurance, reinsurers, insurers, insurance, securities, investors, investment, finance, trading, pension\\
NORTH &\textbf{Education and achievements}: school, schools, graduated, college, students, undergraduate, graduation, graduate, attended, education\\
AMERICA &\textbf{Nike shoes and fashion}: nike, shoes, sneakers, jordans, jeans, tops, black, boys, men, casual\\
&\textbf{Family dynamics and relationships}: family, families, children, kids, grandchildren, relatives, grandparents, parents, child, parent\\
&\textbf{Cape Cod news}: lifeguard, drowned, drowns, newstweet, hospitalized, snorkeling, cape, reported, reuterstweet, pulled\\
&\textbf{Reading and fiction}: books, book, reading, novels, series, enjoyed, novel, romance, katniss, readers\\
&\textbf{Male terms}: male, man, boy, brother, he, him, his, son, Liam, Noah, Ethan, Jacob, William, Michael, James, Alexander, Benjamin, Matthew\\
&\textbf{Female terms}: sister, mother, aunt, grandmother, daughter, she, hers, her, Emma, Olivia, Ava, Sophia, Isabella, Mia, Charlotte, Amelia, Harper, Evelyn\\

\midrule
 & \textbf{Religious beliefs and figures}: god, gods, bible, mankind, faith, christ, spiritual, christian, religion, jesus\\
&\textbf{Family relationships}: mum, mother, mom, mums, parent, family, parents, baby, dad, father\\
&\textbf{Music record and Artists}: music, album, albums, jazz, songs, hits, musicians, artists, recordings, blues\\
&\textbf{Woordworking plans and projects}: plans, furniture, woodwork, wood, woodcraft, woodworking, plywood, carpentry, cabinets, wooden\\
&\textbf{Building and designing boats}: boatbuilder, boatbuilding, boats, plans, boat, sauceboat, sailboat, build, catamaran, kits\\
&\textbf{Weight loss and nutrition for women}: diet, workout, exercise, foods, weight, food, eating, healthy, pounds, fat\\
OCEANIA&\textbf{Superheroes and their Universes}: superhero, superheroes, avengers, marvel, comics, superman, aquaman, heroes, comic, hero\\
&\textbf{Exercises for hormone development}: hormones, weightlifting, workouts, deadlifts, hormonal, exercises, lifting, testosterone, fitness, squats\\
&\textbf{Building and designing boats}: boatbuilder, boatbuilding, boats, plans, boat, sauceboat, sailboat, build, catamaran, kits\\
&\textbf{Kids furniture and decor}: furniture, chairs, sofas, ikea, sofa, cushions, sectional, upholstered, couch, childrens\\
&\textbf{Male terms}: male, man, boy, brother, he, him, his, son, Manaia, Tane, Kai, Ariki, Mika, Koa, Rangi, Kane, Tama, Hemi\\
&\textbf{Female terms}: sister, mother, aunt, grandmother, daughter, she, hers, her, Aroha, Moana, Tui, Lani, Kahurangi, Ariana, Malie, Marama, Ava, Kaia\\
\bottomrule  
\end{tabular}  
\caption{\label{tab:weat_list}
Word lists corresponding to each topic for computing region-aware WEAT metric}
\end{table*}

\section{Paired-list for F-M datasets}
\label{bolukbasilist}
Here is the list of the 52 pairs used to create the F-M datasets per region: \newline
[monastery, convent], [spokesman, spokeswoman], [Catholic priest, nun], [Dad, Mom], [Men, Women], [councilman, councilwoman], [grandpa, grandma], [grandsons, granddaughters], [prostate cancer, ovarian cancer], [testosterone, estrogen], [uncle, aunt], [wives, husbands], [Father, Mother], [Grandpa, Grandma], [He, She], [boy, girl], [boys, girls], [brother, sister], [brothers, sisters], [businessman, businesswoman], [chairman, chairwoman], [colt, filly], [congressman, congresswoman], [dad, mom], [dads, moms], [dudes, gals], [ex girlfriend, ex boyfriend], [father, mother], [fatherhood, motherhood], [fathers, mothers], [fella, granny], [fraternity, sorority], [gelding, mare], [gentleman, lady], [gentlemen, ladies], [grandfather, grandmother], [grandson, granddaughter], [he, she], [himself, herself], [his, her], [king, queen], [kings, queens], [male, female], [males, females], [man, woman], [men, women], [nephew, niece], [prince, princess], [schoolboy, schoolgirl], [son, daughter], [sons, daughters], [twin brother, twin sister]. \newline
Each pair in the above is denoted as a [male, female] pair.

\section{Llama 2 prompt for topic modeling}

\label{prompt}
The prompt scheme for \texttt{Llama2} consists of three prompts: (1) System Prompt: a general prompt that describes information given to all conversations, (2) Example Prompt: an example that demonstrates the output we are looking for, and (3) Main Prompt: describes the structure of the main question, that is with a given set of documents and keywords, we ask the model to create a short label for the topic. Fig~\ref{fig:prompt} displays the three prompts as used in the code.

\section{Topic Cluster Labels using other LLMs}

\label{sec:gpt_topics}

We use \texttt{Llama2} to fine-tune our topics to label them for better coherence in our paper. However, we also experiment with \texttt{GPT-4} and arrive at similar topics in Table~\ref{gpt_topics}. (see Table~\ref{continent_toptopics} for comparison with \texttt{Llama2} topic labels).

\section{Topic Word Clusters Example - Africa}
\label{sec:topic_example}
\begin{figure*}
    \centering
    \includegraphics[width=1\linewidth]{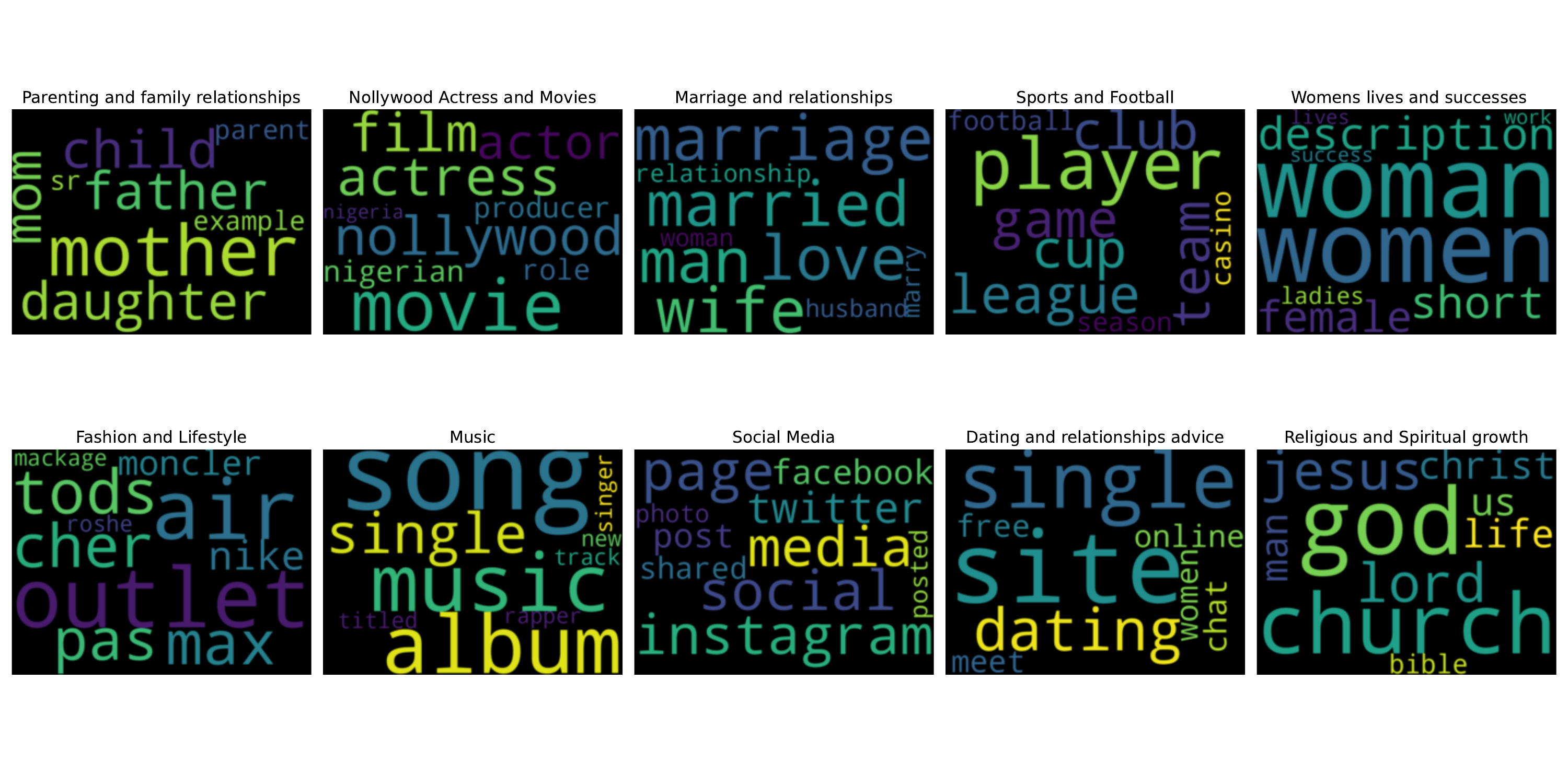}
    \caption{\centering Topic Word Clusters - Africa}
    \label{fig:topic_africa}
\end{figure*}

Here, we provide an example of how topics look in our data. In Fig~\ref{fig:topic_africa}, we provide word clusters of topics from Africa. The word clusters contain the top 10 words from each topic in Africa. We find that topic labels by \texttt{Llama2} are coherent in terms of top topic words. 

\section{Region specific BERTs to identify top words in F/M direction}
\label{top_appendix}

\begin{figure*}
\centering
  \begin{subfigure}{0.55\textwidth}
    \includegraphics[width=\textwidth]{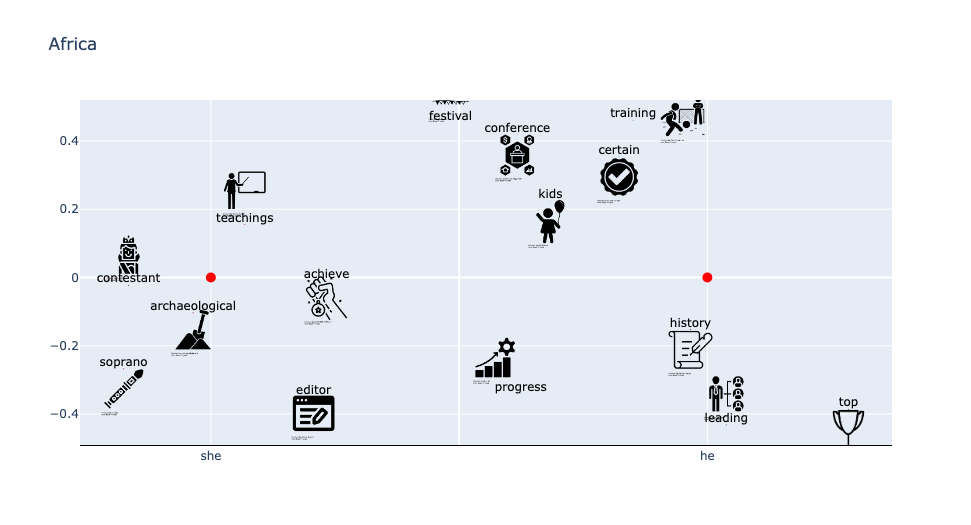}
  \end{subfigure}  
  \begin{subfigure}{0.55\textwidth}
    \includegraphics[width=\textwidth]{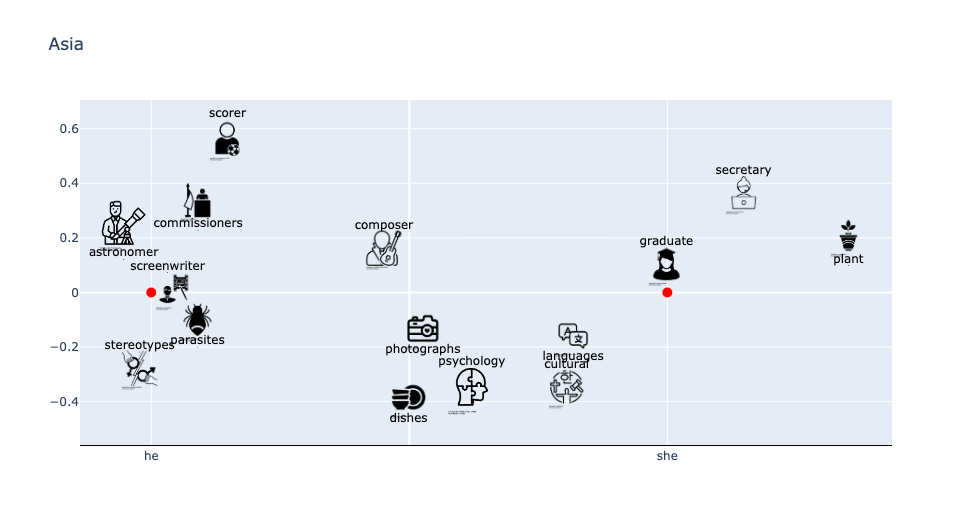} 
  \end{subfigure}  
  \begin{subfigure}{0.55\textwidth}
    \includegraphics[width=\textwidth]{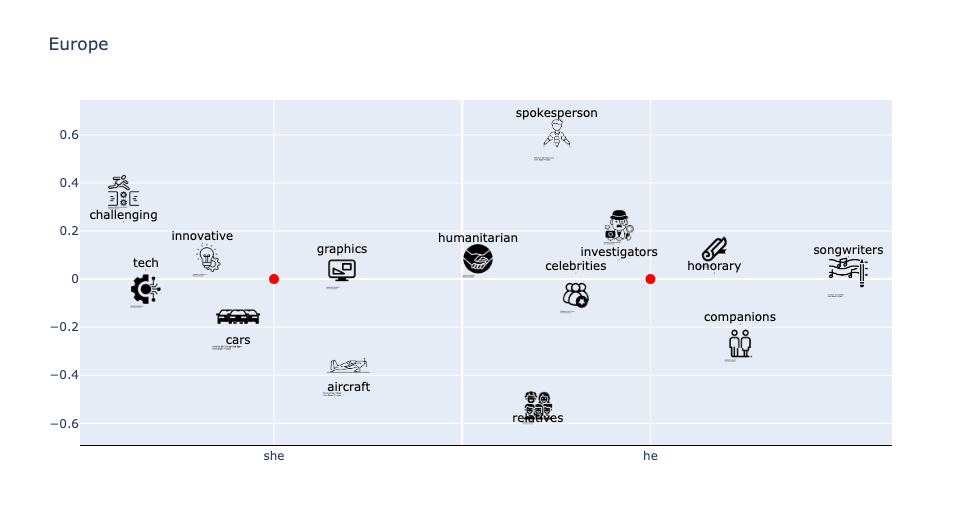}  
    \label{fig:image3}  
  \end{subfigure}  
  \begin{subfigure}{0.55\textwidth}
    \includegraphics[width=\textwidth]{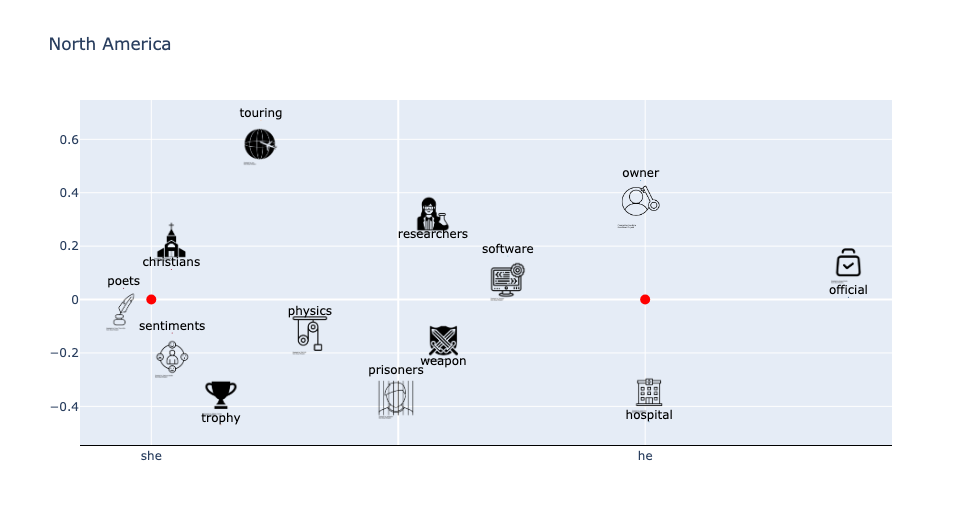}  
  \end{subfigure}  
  \begin{subfigure}{0.55\textwidth}
    \includegraphics[width=\textwidth]{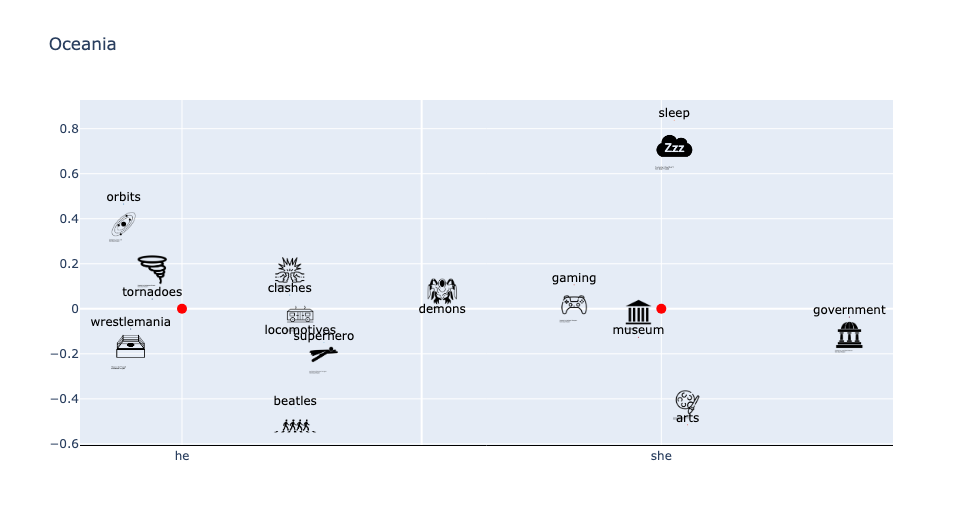}  
  \end{subfigure}  
  \caption{\centering Top words for each region(Africa, Asia, Europe, North America and Oceania) using region-specific BERTs}  
  \label{fig:top}  
\end{figure*}

To motivate our case to investigate differences in biases across regions, we use BERT to compute the top words corresponding to the \textit{she-he} axis in the embedding space. BERT is a pre-trained transformer-based language model that consists of a set of encoders. As a motivation experiment to identify differences in the contextual embedding space for different regions, we fine-tune BERT with the masked language modeling task (no labels) for each region separately. For a given word, we compute its embeddings by averaging out all sentence embeddings where it occurs across the dataset.Similarly, we compute embeddings for all words in the dataset. The tokenized input goes through the BERT model and we take the hidden states at the end of the last encoder layer (in our case, BERT-base, i.e. 12 encoder layers) as sentence embeddings. We identify the top words with the highest projection across the \textit{she-he} axis in the region-specific datasets. If we find differences in the top words across regions, it is possible that dominating bias topics vary by region as well. Fig~\ref{fig:top} shows the top words closest to `she' and `he' contextual embeddings in our data for each region. We find that top words differ quite a bit across different regions. 
We find many differences in the top F (close to $she$) and M (close to $he$) words across regions. Some top F words are soprano, archaeological (Africa); graduate, secretary (Asia); innovative, graphics (Europe); poets, sentiments (NA); and arts, sleep (Oceania). Some top M words are history, leading (Africa); astronomer, commissioners (Asia); honorary, songwriters (Europe); owner, hospital (NA); and wrestlemania, orbits (Oceania). Gender-neutral words such as poets, secretaries, astronomers, commissioners, songwriters, owners, and so on are closer to either the she or he axes. Although comparable to the findings of \cite{bolukbasi2016man}, the variances among regions inspire us to look deeper into the data to arrive at culture-specific bias themes.

\section{Implementations details}
\label{hyperparam}

For training our \texttt{Bertopic} model, we use \texttt{Google Colab's} Tesla T4 GPU, and it takes 15 min to run topic modeling for a region-specific F-M dataset. Region-specific BERTs are run on NVIDIA RTX2080 GPUs. Each BERT training experiment takes 1 GPU hour. For our LLM experiment, we used NVIDIA-A40 for \texttt{Mistral-7b-Instruct} and \texttt{Llama-3-8b} for an hour. We do not use any GPUs for \texttt{GPT-4}, \texttt{Claude-3-Sonnet} and \texttt{Gemini-Pro}. 

\subsection{Bertopic}
\label{bertopic}

We use Bertopic's default models: \texttt{SBERT} \cite{reimers-2019-sentence-bert} to contextually embed the dataset, \texttt{UMAP} \cite{mcinnes2020umap} to perform dimensionality reduction, \texttt{HDBSCAN} \cite{Malzer_2020} for clustering to perform topic modeling. We choose the embedding model \texttt{BAAI/bge-small-en} from \emph{Huggingface} \cite{hugg}. We set \texttt{top\_n\_words} to 10 and \texttt{verbose} as True and set the \texttt{min\_topic\_size} to 100 for the \texttt{Bertopic} model. Finally, we use Bertopic's official library to implement the model. 

\subsection{Llama2}
\label{Llama2}
We use \texttt{Llama2} to finetune the topics to give shorter labels for each topic. We set the \texttt{temperature} to $0.1$, \texttt{max\_new\_tokens} to $500$ and \texttt{repetition\_penalty} to $1.1$. We utilize Bertopic's built-in representation models to use \texttt{Llama2} in our topic model. 

\subsection{LLM experiment}
For \texttt{GPT-4}, and \texttt{Mistral-7b-Instruct} and \texttt{Llama-3-8b}, we utilize the Microsoft Azure API\footnote{\url{https://learn.microsoft.com/en-us/rest/api/azure/}}, huggingface\footnote{\url{https://huggingface.co/mistralai/Mistral-7B-Instruct-v0.1}}, and huggingface\footnote{\url{https://huggingface.co/meta-llama/Meta-Llama-3-8B}} for inference respectively. We use a temperature $0.8$ for all models. For \texttt{Gemini-Pro} and \texttt{Claude-3-Sonnet}, we use the available chat interface.

\subsection{Region-specific BERT}
We use the uncased version BERT \cite{devlin-etal-2019-bert} for our region-specific BERT model trained for the MLM objective. We use a batch size of 8, a learning rate of $1 \cdot 10^{-4}$, and an AdamW optimizer to train our BERT models for 3 epochs.

\section{Human Validation}
\label{sec:screenshots}

\begin{figure*}
    \centering
    \begin{subfigure}{\textwidth}
        \centering
        \includegraphics[width=0.5\textwidth]{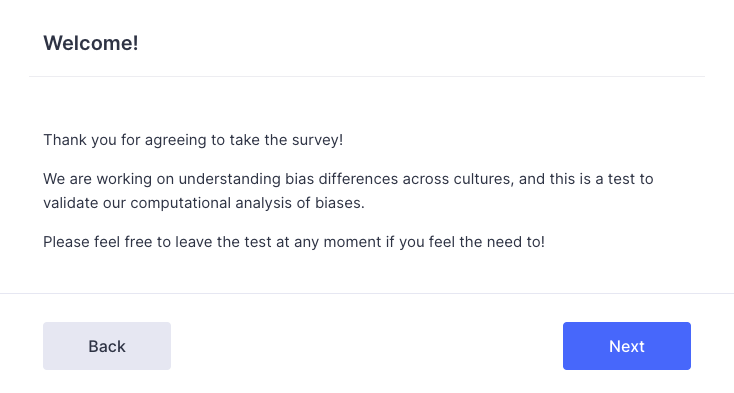}
    \end{subfigure}
    \begin{subfigure}{\textwidth}
        \centering
        \includegraphics[width=0.5\textwidth]{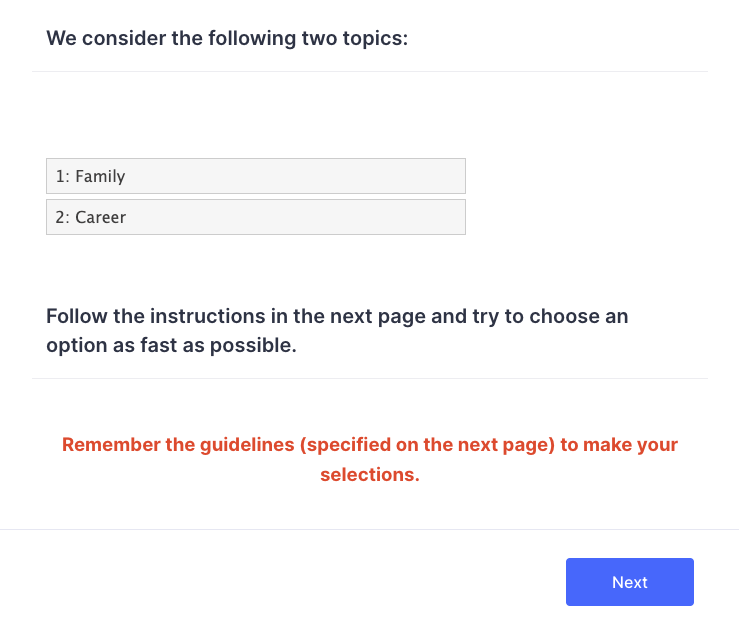}
    \end{subfigure}
    \\
    \begin{subfigure}{\textwidth}
        \centering
        \includegraphics[width=0.5\textwidth]{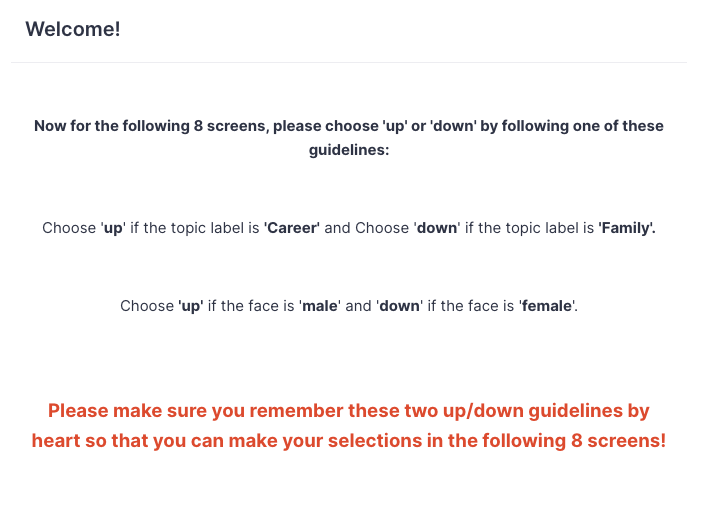}
    \end{subfigure}
    \\
    \begin{subfigure}{\textwidth}
        \centering
        \includegraphics[width=0.5\textwidth]{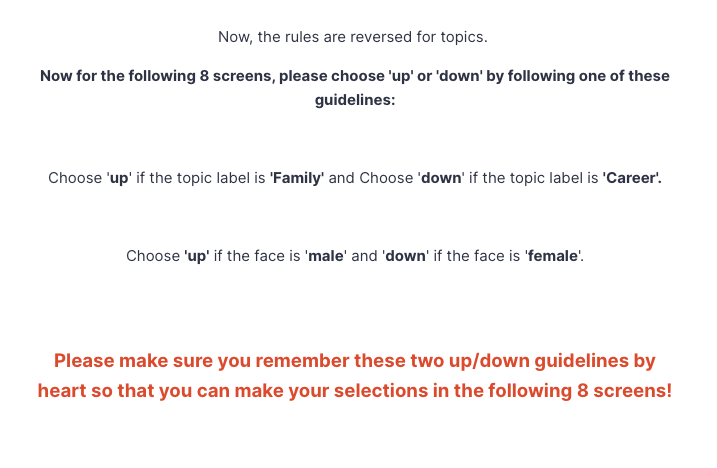}
    \end{subfigure}\\
        \begin{subfigure}{\textwidth}
        \centering
        \includegraphics[width=0.5\textwidth]{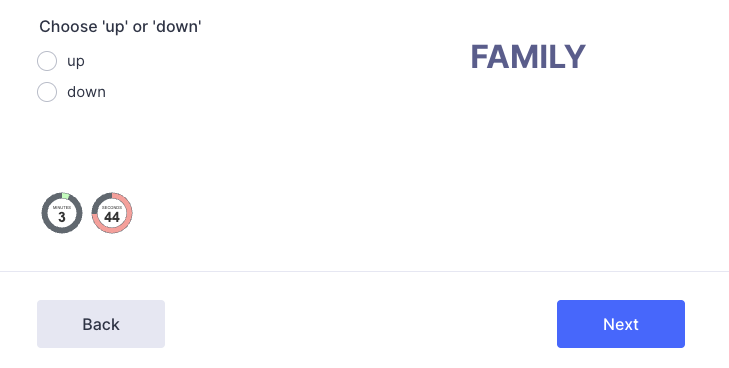}
    \end{subfigure}
    \\

    \caption{\small \centering Annotation Form Screenshots (We do not include screenshots with faces to protect privacy)}
    \label{fig:forms}
\end{figure*}

Students and staff from a college campus were recruited as annotators in the study. Screenshots of the form are displayed in Fig~\ref{fig:forms}. We have 6 annotators per region (3 male and 3 female).

\section{Reproducibility}
\label{sec:repro}
We open-source our codes, which are uploaded to the submission system. We include commands with hyperparameters in our codes. This would help future work to reproduce our results.

\end{document}